\RequirePackage{fix-cm}
\documentclass[<smallextended,final>]{svjour3}
\smartqed

\usepackage[pdfstartview=FitH,
            CJKbookmarks=true,
            bookmarksnumbered=true,
            bookmarksopen=true,
            colorlinks,
            pdfborder=001,
            linkcolor=green,
            anchorcolor=green,
            citecolor=green
            ]{hyperref}
\usepackage{graphicx}
\usepackage{multirow}
\usepackage[cmex10]{amsmath}
\usepackage{url}
\usepackage{enumerate}
\usepackage{subfigure}
\usepackage{amsfonts}
\usepackage[table]{xcolor}
\usepackage{hhline}
\usepackage{cite}

\graphicspath{{images/expt/}{images/body/}}

\begin{document}

\title{Indexing of the CNN Features for the Large Scale Image Search}

\author{Ruoyu Liu \and Shikui Wei \and Yao Zhao \and Yi Yang}

\institute{Ruoyu Liu, Shikui Wei and Yao Zhao \at Beijing Jiaotong University
           \and
           Yi Yang \at University of Technology Sydney
}

\date{Received: date / Accepted: date}

\maketitle

\begin{abstract}
The convolutional neural network (CNN) features can give a good description of image content, which usually represent images with unique global vectors. Although they are compact compared to local descriptors, they still cannot efficiently deal with large-scale image retrieval due to the cost of the linear incremental computation and storage. To address this issue, we build a simple but effective indexing framework based on inverted table, which significantly decreases both the search time and memory usage. In addition, several strategies are fully investigated under an indexing framework to adapt it to CNN features and compensate for quantization errors. First, we use multiple assignment for the query and database images to increase the probability of relevant images' co-existing in the same Voronoi cells obtained via the clustering algorithm. Then, we introduce embedding codes to further improve precision by removing false matches during a search. We demonstrate that by using hashing schemes to calculate the embedding codes and by changing the ranking rule, indexing framework speeds can be greatly improved. Extensive experiments conducted on several unsupervised and supervised benchmarks support these results and the superiority of the proposed indexing framework. We also provide a fair comparison between the popular CNN features.
\end{abstract}

\section{Introduction}

Image search aims to find relevant images of a specific query from mass data, which is a fundamental problem in many real-world scenarios. In the last two decades, the main effort focus has been on improving both search accuracy and efficiency. For example, Yang \emph{et al.} have proposed a semi-supervised ranking algorithm for image retrieval \cite{yang2012multimedia}. Search accuracy is closely related to feature extraction, whereas search efficiency is mainly dependent on indexing structure.

To represent image content accurately, various feature extraction schemes have been developed \cite{colorhistogram_han2002fuzzy,tamura1978textural,hu1962visual,sift_lowe2004distinctive,gloh_mikolajczyk2005performance,surf_bay2006surf,hog_dalal2005histograms}, which can be roughly separated into two groups: global features and local features. Global features such as color histograms \cite{colorhistogram_han2002fuzzy}, Tamura texture features \cite{tamura1978textural} and moment invariants \cite{hu1962visual}, are often statistics of image color, textural or shape information, where each image is described as a single short vector. Generally, this kind of feature is compact and efficient for performing image searches. However, they cannot describe image details or handle some transformations like rotation or changes in illumination, since they only capture low-level information. By contrast, local features, \emph{i.e.}, SIFT \cite{sift_lowe2004distinctive}, GLOH \cite{gloh_mikolajczyk2005performance}, SURF \cite{surf_bay2006surf} and HOG \cite{hog_dalal2005histograms}, describe images with a set of local descriptors and are better at discriminating between content. However, their shortcoming is a huge amount of local features, even when extracted from a small-scale dataset. When facing a large-scale collection, image searches based on local features will result in unacceptable computation and storage costs. A summary of global and local features can be found in~\cite{liu2007survey} and \cite{li2015survey} respectively. Another branch of global features aggregates the local descriptors of an image (for example, SIFT) into a unique vector. Two famous such features are Fisher vectors \cite{perronnin2007fisher} and VLAD \cite{jegou2010aggregating}. Together with a dimension reduction technique (\emph{e.g.}, PCA \cite{jolliffe2002principal}), they offer a good solution for large-scale image retrieval. Recent works have tried to increase VLAD's performance \cite{gong2014multi,gaodemocratic,dong2016holons} and use it to aggregate CNN vectors~\cite{liu2015uniting}.

To speed up searching, various indexing schemes have been studied. Most tree-based structures, such as R-tree \cite{r-tree_guttman1984r}, M-tree \cite{m-tree_patella1997m} and inverted table \cite{ivtable-sivic2003video,zhong2015fast,zheng2015fast}, generally divide the entire feature space into several non-overlapping partitions. The query image simply scans the partitions near its location during retrieval. Since only a small proportion of the database images are compared to the query image, search efficiency is significantly improved. Another group of indexing schemes are hashing methods \cite{hashing_datar2004locality,weiss2009spectral,zhang2010self,wang2013order,ji2012super,zhou2014towards,liu2014contextual,wang2012semi,lv2015asymmetric}, which learn a set of hash functions to map images from their original feature space into a binary space. Since fast bitwise operations (\emph{i.e.}, XOR) are employed to calculate Hamming distance, the search speed is boosted greatly. A good survey of hashing methods can be found in \cite{wang2016survey}.

While many image indexing frameworks have been proposed, inverted tables \cite{ivtable-sivic2003video} are still an effective structure, since their search time and memory usage are much lower than using individual descriptors directly \cite{jegou2008hamming}. Previous versions of inverted tables were designed for a bag-of-words (BoW) model \cite{ivtable-sivic2003video} (also named a bag-of-features in \cite{jegou2010improving}), where an image is represented by a set of local features. However, local features are not suitable for large-scale retrieval, because they are huge even when extracted from a small or medium collection. When the database contains numerous images, computation and storage costs quickly increase. Recently, feature learning techniques based on convolutional neural networks (CNNs) \cite{razavian2014cnn} have become popular. Instead of representing images with human-designed descriptors, these schemes encode each image into unique global vectors with a deep network, and also provide a high-level description of the image's content \cite{babenko2014neural}. For these features, hashing seems like a good choice for an indexing structure. However, it still fails to overcome the large-scale issue, since the search time increases with the increase in database volume. Therefore, building an efficient indexing framework for these global features, especially for large-scale image retrieval, is necessary.

\begin{figure}[t]
\begin{center}
   \includegraphics[width=0.98\linewidth]{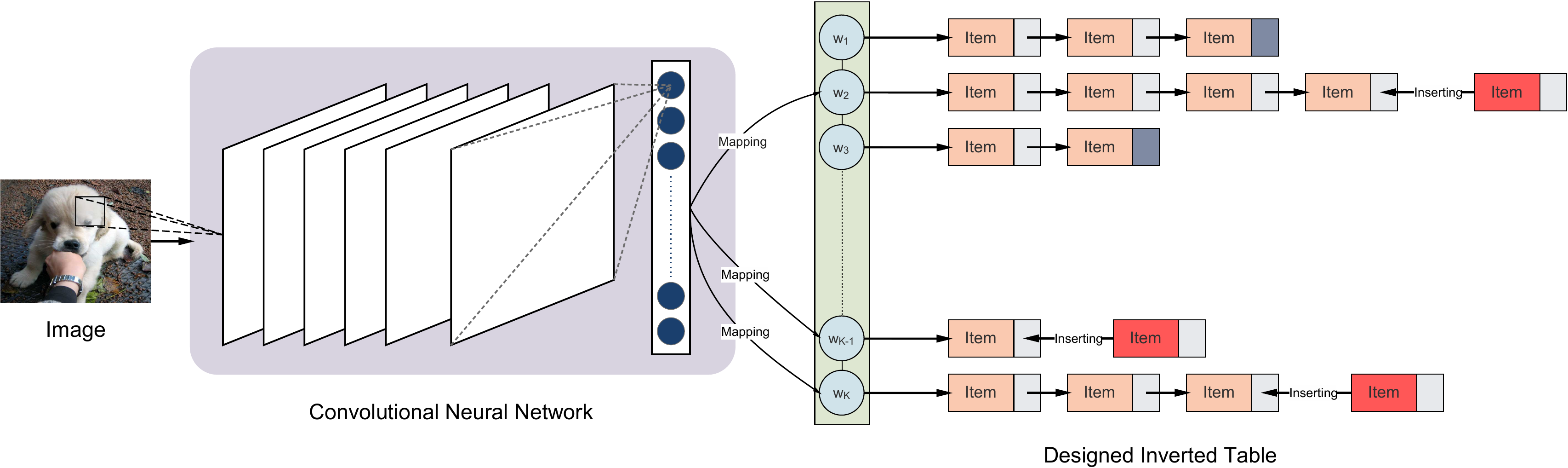}
\end{center}
   \caption{Illustration of the basic principles of our proposed method. Each image is encoded by a pre-trained convolutional neural network. Then, the extracted feature vector is mapped to multiple codewords of an inverted table and its items are inserted into the corresponding lists.}
\label{fig:BasicIdea}
\end{figure}

In this paper, we attempt to design an effective indexing structure for CNN features. Our approach draws inspiration from inverted tables, which builds similar structures for the descriptors. The basic principles of the proposed method are illustrated in Fig. \ref{fig:BasicIdea}, which contains two main procedures. First, images are encoded by a pre-trained CNN into feature vectors. Then, these global features are properly inserted into the proposed inverted framework. The benefits, that make the design viable, are two-fold. First, the feature space is divided into several Voronoi cells by constructing a visual dictionary, then each query only needs to compare a few candidates located in its nearby regions which reduces calculations to greatly improve search speed. Second, the descriptors are compressed when stored in an inverted table, which vastly reduces the storage required. However, previous schemes were designed for local features, so some modifications are required to accommodate CNNs. We summarize two key problems to be solved: (1) How to build a proper inverted table for CNNs, where the key lies in codebook construction and vector quantization. (2) How to compensate for quantization error since it seriously impacts efficacy when images are represented by single vectors. Our solutions are briefly introduced in the following summary, and we give detailed explanations in Section \ref{sec:OverviewOfTheIndexingFramework}.

The main contributions of this paper are summarized as follows:
\begin{enumerate}[(1)]
\item We propose a simple but effective indexing technique for CNNs based on inverted tables. For a specific image collection, the number of global features extracted is much lower than local descriptors. Considering this issue, we employ a partitioned k-means (PKM) \cite{wei2013partitioned} to construct a visual dictionary with limited samples. Vector quantization is performed using a simple nearest neighbor search. We demonstrate that both computation and storage costs can be significantly reduced with this simple structure. To the best of our knowledge, no similar indexing structures have been published.

\item We use several auxiliary strategies to compensate for quantization errors, since this is the main impediment to effectiveness when an image is represented by a single CNN descriptor. More concretely, we use linear segment embedding (LSE) \cite{kernel_wei2013joint} to calculate a compact representation for each vector, which improves precision by pruning false matches. In addition, we use multiple assignment (MA) \cite{jegou2007contextual} for both query and database images to increase the probability of true matches existing in the candidate set. Interestingly, when we replace LSE with a hashing scheme and change the ranking rule, the search speed is further boosted. We demonstrate that this small change makes the proposed indexing more efficient, especially in large-scale scenarios.

\item We fairly compare the performance of several popular CNNs under our indexing framework (we provide a short review of these networks in Section \ref{sec:RelatedWork}). These CNNs have different depths, layer structures and most importantly description capabilities in image content, but they all encode each image into a fixed-length vector. We demonstrate that our proposed indexing is widely suitable for various CNNs, and significantly improve search speeds with little loss of precision.
\end{enumerate}

The rest of the paper is organized as follows. Fist, we review the previous work on CNNs and inverted tables. Then, we provide an overview of the proposed indexing framework and its details in Sections \ref{sec:OverviewOfTheIndexingFramework} and \ref{sec:detail-of-proposed-framework}. The experimental results and conclusions are given in the last two sections.

\section{Related Work}
\label{sec:RelatedWork}

In this section, we provide a short review of popular convolutional neural networks, then introduce inverted tables and their related techniques. Discussing these basic works helps to understand our proposed indexing framework.

\subsection{Convolutional Neural Networks}
\label{subsec:CNN}

CNNs have drawn considerable attention from both academic and industry communities due to their excellent ability to represent image content \cite{babenko2014neural,razavian2014cnn,wei2014cnn}. Yan \emph{et al.} have used CNN feature to represent images for classification \cite{yan2016image}. Zheng \emph{et al.} have proposed a fusion scheme to fuse CNN and other image features \cite{zheng2015query}. They are a branch of deep neural networks (DNNs), and are commonly used to encode image content. The landmark work on DNNs is \cite{hinton2006reducing}, which focuses on dimension reduction. In \cite{hinton2006reducing}, the two foundations of DNN were introduced, \emph{i.e.}, initialization and fine-tuning, and a generative model, named deep belief network, was proposed which gave a better digit classification than the best discriminative learning algorithms of the time. Krizhevsky \emph{et al.} won the \textit{ILSVRC-2012} competition with an eight-layer CNN named AlexNet \cite{cnn_krizhevsky2012imagenet}. AlexNet's high classification capability brought significant research interest to CNNs, and as a result many CNNs were subsequently proposed.

All CNNs have similar structures, \emph{i.e.}, combinations of multiple convolutional and fully connected layers. LeNet-5 \cite{LeNet-5_lecun1998gradient} is a classic early CNN with $ 7 $ layers. Despite its simplicity, it has achieved great success rate in recognizing handwritten digits. Network in network \cite{NIN_lin2013network} is a recent CNN, which builds micro networks in place of linear filters to improve the abstraction of receptive fields. This idea was extended by GoogLeNet \cite{GoogLeNet_szegedy2015going}, which used a $ 22 $-layer network to win the \textit{ILSVRC-2014} competition. GoogLeNet uses well-designed inception modules as layers, which allows the network to increase the unit number without increasing computational complexity. Recently, there has been a tendency to continuously grow the depth of CNNs. DeepID \cite{sun2014deep} has $ 9 $ layers, and VGG nets \cite{VGG_simonyan2014very} have $ 16 $ or $ 19 $ layers. The champion of the \textit{MS COCO 2015} competition scheme, deep residual network~\cite{he2015deep}, has a depth that reaches $ 152 $ layers. Meanwhile, it has been shown that the top fully connected layers of CNNs provide high-level descriptions of image content \cite{babenko2014neural}.
 
Recent works have also employed CNNs to detect and describe the objects inside images, for example, R-CNN \cite{RCNN_girshick2014rich,FastRCNN_girshick2015fast,FasterRCNN_ren2015faster}, SPP net \cite{SPPnet_he2015spatial} and YOLO \cite{YOLO_redmon2015you}. In these networks, the features are still global, since they are the descriptors of the regions in an image (which are big enough to cover the objects), but they only number a handful (which is relevant to the number of objects). 

Although CNN features are more compact than local descriptors (usually each image is represented by a single 4096D or 1000D vector), they still face the issue of scale, since image quantity is increasing so fast. A brute-force search of original features or hash codes is not a good solution, since search time increases linearly according to database volume, and becomes unacceptable after the number of images reaches a critical value. Therefore, it is necessary to build effective indexing for CNNs. 

\subsection{Inverted Table}
\label{subsec:InvertedTable}

Inverted tables were initially developed to index documents for text retrieval \cite{baeza1999modern}. To facilitate image matching on local features, Sivic \emph{et al.} \cite{ivtable-sivic2003video} extended this technique to indexing images via a BoW model \cite{nister2006scalable,nowak2006sampling}. The principle behind BoW is to construct a visual dictionary and represent each image as an orderless collection of words -- a text document. An inverted table can then be directly to each image as a text document. Fig. \ref{fig:BasicIdea} illustrates the basic structure of inverted tables, which contain a fixed-sized visual dictionary. Each local feature makes an item, and it is inserted into the corresponding list of its nearest visual word. Each item contains the image's ID and an embedding code, where the latter helps to compensate for the quantization error between the feature and visual word. In the search phase, a voting process is performed to measure the similarity between query image and database images.

Much research to improve the performance of inverted tables has been undertaken. To effectively construct a visual dictionary, some fast clustering methods have been proposed to lower the time cost of building the dictionary \cite{nister2006scalable,PQ_jegou2011product,wei2013partitioned}. Product quantization (PQ) \cite{PQ_jegou2011product} is a quantization approach that approximates nearest neighbor searches which map a vector to a codeword. In this way, the distance between two vectors is estimated by the distance between two codewords. PQ decomposes the feature space into a Cartesian product of low-dimensional subspaces and quantizes the vectors in each subspace separately. It can easily generate a codebook with an exponentially large number of code words. Inspired by PQ, PKM \cite{wei2013partitioned} splits the full space into a set of subspaces and then clusters the samples separately. In this way, PKM can build a large and unbiased visual vocabulary using a small training set. Other schemes, such as \cite{ge2014optimized,kalantidis2014locally,SPQ_ning2016scalable}, mainly focus on improving PQ's performance. 

In addition, Hamming embedding \cite{jegou2008hamming,kernel_wei2013joint} has been proposed to improve the search accuracy of inverted tables. These schemes generate a short binary code for each feature, which is included in the item of inverted tables together with the image ID. Since search accuracy is mainly impacted by quantization errors, Hamming embedding tries to remove false matches to improve precision. In the search phase, the items whose distances are higher than a particular threshold drop out of voting. Although Hamming embedding and hashing are similar, as they both compress the feature vectors into binary codes, they are totally different in application. Hamming embedding is a technique to assist inverted tables, while the hashing is used to quickly calculate approximate distances between features.

Compared to inverted tables, hashing seems a better solution for indexing CNN features, since it is perfectly suited to global descriptors. However, a linear increase in search time restricts its application and makes it inefficient for searches in a large-scale databases. However, inverted tables could solve this issue by partitioning the feature space into Voronoi regions (when a visual dictionary is built), and only a handful of images are really compared, whose number can be flexibly controlled by codebook size. Despite big success in the BoW model, CNN features have not been able to benefit from this approach, since it was initially designed for local descriptors. Therefore, it is necessary to reform inverted tables for CNNs. Moreover, our second scheme draws inspiration from both hashing and inverted tables.

\section{Overview of the Indexing Framework}
\label{sec:OverviewOfTheIndexingFramework}

To better understand the proposed indexing framework, we give a brief review of the key problems and their solutions before we introduce the details.

\subsection{Problem Description}
\label{subsec:ProblemDescription}

Suppose there is a set of images encoded by CNNs, our goal is to develop effective indexing to perform fast image searches with little loss of precision. An optional solution is indexing by hashing. However, its search time linearly increases with image quantity, which is inefficient when the database is massive. To address large-scale searches, it is necessary to develop better frameworks for CNN features. Inspired by inverted tables, which partition the feature space to only compare a small proportion of the images, we design a similar indexing framework to reform the original structure to match the CNN vectors. We summarize the two main problems to be addressed in the following.\\

\noindent \textbf{Problem $1$:} How to modify inverted table to make it suitable for global CNN vectors?\\

In traditional inverted tables, images are indexed by inserting local features into corresponding lists of visual words according to their quantization results. That is, this framework is tailored for local features. Because CNNs generate global descriptors, we need to modify the original structure of the inverted table to adapt it for CNN vectors.\\

\noindent \textbf{Problem $2$:} How to compensate for quantization errors, since they are a major contributor to inefficiency when quantizing CNN features into visual words?\\

When we take visual words to approximately represent global features, quantization errors between the original and approximate representations occur. Particularly, when images are represented by unique vectors (\emph{e.g.}, CNN features), these errors seriously impact precision, since dissimilar images may be quantized into the same visual word and vice versa. Auxiliary compensation strategies are therefore needed to improve search results.

\subsection{Solution Overview}
\label{subsec:SolutionOverview}

The two problems mentioned above are not independent, and the solution to Problem 2 can be found in solving Problem 1. However, for clarity, we provide solutions to these problems in sequence.

To solve Problem $1$, we make a simple modification to the original structure of inverted tables. For the same image set, since the amount of CNN (global) features is much smaller than local descriptors, we perform PKM to construct a large visual dictionary using limited samples. Then, each vector generates items as usual which are inserted into the inverted table later. The other modification can be found in the solution to Problem 2.

To solve Problem $2$, we employ several strategies to compensate for quantization errors. The ill effect of the quantization error is reflected in the loss of true matches, because dissimilar images may be located in the same Voronoi regions and vice versa. Since each image is represented as a single vector, if it is only quantized into the nearest visual word, the probability of finding positive results is relatively small. To increase this probability, we use multiple assignment (MA) \cite{jegou2010improving} for both the query and database images, where each vector is mapped to multiple nearest visual words. In this way, there is an increased chance to find positive results in the candidate set. Additionally, Hamming embedding (\emph{i.e.}, LSE \cite{wei2013partitioned}) is also introduced to help further improve precision. Interestingly, we can also use hashing (\emph{i.e.}, LSH \cite{hashing_datar2004locality} and VDSH \cite{VDSH_zhang2016efficient}) in place of Hamming embedding to further improve efficiency. In essence, when MA is employed, images are represented in a similar way to BoW, so we can use inverted tables to index images. The key difference between our devised framework and traditional inverted tables (including auxiliary schemes) is that ours focuses on global features, while the latter is tailored to local descriptors. 

\section{Details of the Proposed Framework}
\label{sec:detail-of-proposed-framework}

In this section, we introduce the details of the proposed framework. First, we describe an initial version that uses the previous techniques in a flexible way to construct an effective inverted table for CNN features. Then, we analyze its bottlenecks and derive an accelerated version by replacing Hamming embedding with hashing.

\subsection{Indexing for CNN Features}
\label{subsec:indexing-for-CNN-features}

Suppose an image set $ \mathcal{X} = \{x_{1},...,x_{N}\} $ contains $ N $ images described by $ D $-dimensional CNN vectors, where $ x_{i} \in \mathbb{R} ^{D}, i=1,...,N $. The final goal of our work is to use inverted tables to perform fast and accurate image retrieval on these CNN vectors. To this end, a visual codebook $ \mathcal{W} = \{w_{1},...,w_{K}\} $ is constructed, where $ K $ is the codebook size. If $ K $ is small, that is, a small codebook is built, the efficiency is deteriorated since too many images exist in the same Voronoi cells partitioned by $ \mathcal{W} $, and it generates a large candidate set. However, the number of CNN (global) vectors are fewer than local features that are extracted from the same image set, so it is hard to build a large codebook. To solve that, we employ a partitioned k-means (PKM) \cite{wei2013partitioned}, which first splits each CNN vector into $ M $ segments of equal length, and then performs k-means clustering on each segmented part separately to construct $ M $ sub-codebooks. After that, all the sub-codebooks are combined via a Cartesian product to generate a large and unbiased codebook \cite{wei2013partitioned}, which is used to build an inverted table.

Consider that each image is represented by a single vector, quantization errors heavily influence search accuracy. That is, similar images may be mapped to different code words and vice versa. If each image is only inserted into the list of its nearest visual word, it has a small chance to find similar images that remain in the list. To overcome this shortcoming, we employ multiple a assignment (MA) strategy, which inserts each image into several lists of its nearest visual words, to increase the possibility of finding positive results.

\begin{figure}[t]
\begin{center}
   \includegraphics[width=0.84\linewidth]{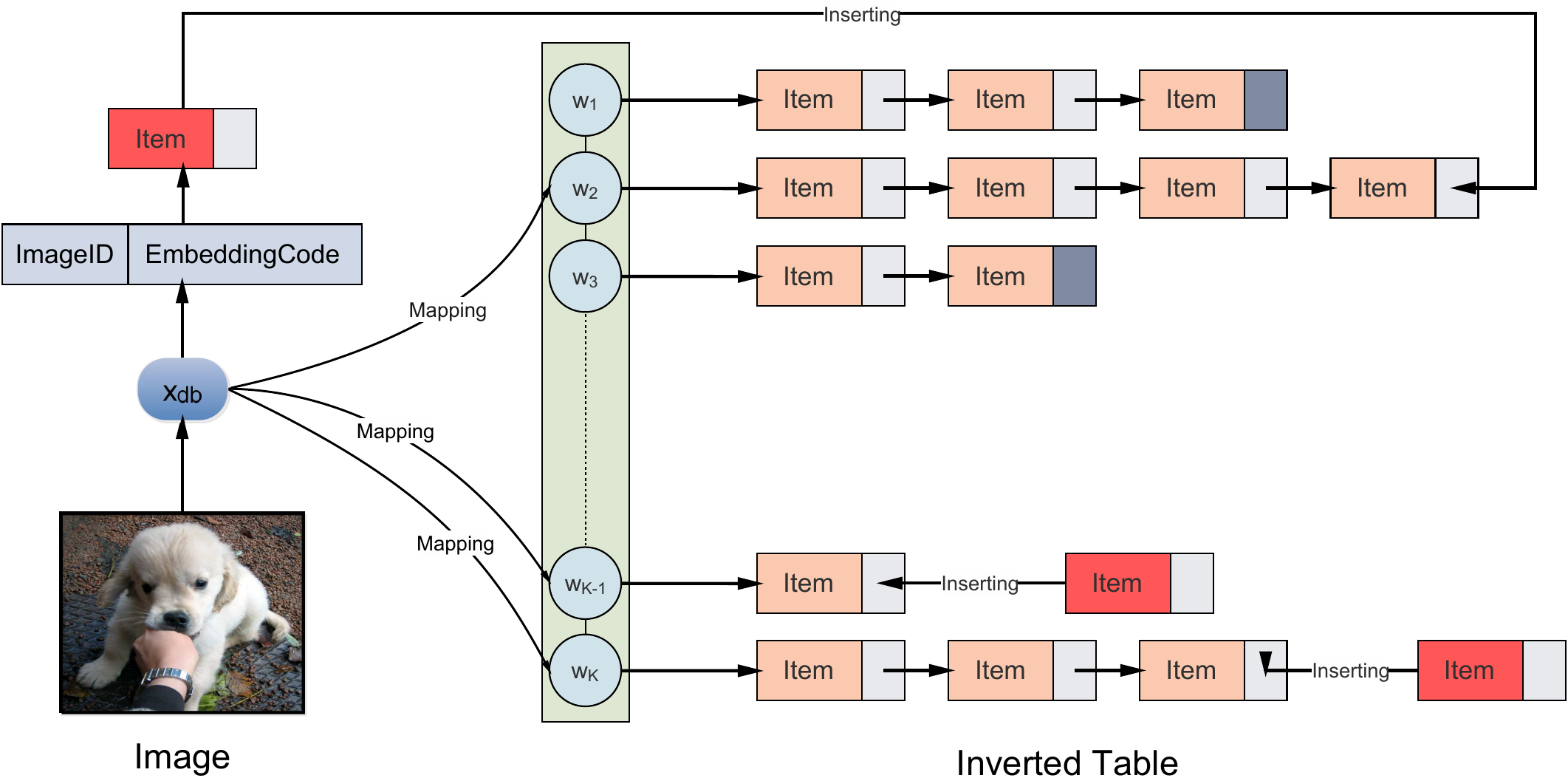}
\end{center}
   \caption{Illustration of $ \mathrm{IFC}_{\mathrm{LSE}} $ indexing structure. Each item of a database feature vector $ X_{db} $ consists the image ID and an embedding code based on the assigned visual word.}
\label{fig:IFCLSE}
\end{figure}

This search scheme is based on voting. Given a query image $ x_{q} $, it first looks into the lists of its $ W $ nearest visual words to build a candidate set with the images inside, where $ W $ is the MA number of query images. Then, the images in the candidate set are sorted by their frequency of occurrence. To further improve search accuracy, Hamming embedding is also introduced into the scheme. We employ linear segment embedding (LSE) \cite{wei2013partitioned}, which calculates $ L $-bit binary codes for each CNN vector based on its mapped visual words. In the search phase, false matches in the candidate set are removed whose distance to the query are beyond te threshold $ T $. The complete indexing structure of the scheme is illustrated in Fig. \ref{fig:IFCLSE}, denoted as $ \mathrm{IFC}_{\mathrm{LSE}} $. The process is summarized as follows: \\

\begin{enumerate}[1.]
\setlength{\itemindent}{1em}
\setlength{\itemsep}{1ex}
\item Construct a large visual dictionary by performing PKM clustering on CNN vectors.
\item Assign each database image to its $ S $ nearest visual words. For each image and one of its assigned words, calculate an LSE code and insert it in the corresponding list with image ID.
\item Each query looks into the lists of its $ W $ nearest visual words and the items whose distances exceed a threshold $ T $ are removed. The images that are left are sorted by their frequency of occurrence and returned.
\end{enumerate}

In practice, we find that $ \mathrm{IFC}_{\mathrm{LSE}} $ achieves high precision when $ S $ and $ W $ equal larger values. However, it loses efficiency in that case and is not superior to hashing. We theorize that the main reason lies in the ranking strategy (\emph{i.e.}, voting) and attempt to design a faster version of the framework to overcome this shortcoming.

\subsection{Faster Indexing for CNN Features}
\label{subsec:faster-indexing-for-cnn-features}

Since the searching of $ \mathrm{IFC}_{\mathrm{LSE}} $ is based on voting, there are two conditions that should be reached to ensure its precision. The first is that positive results should exist in the candidate set, and the second is that their frequency of occurrence should be the highest. Both conditions require the enlargement of the MA numbers ($ S $ and $ W $), which increase the probability of mapping each query and its positive results into the same Voronoi cell. However, since the candidate set is also expanded, the scheme also loses efficiency and is even slower than hashing. In order to design a faster scheme, we first analyze $ \mathrm{IFC}_{\mathrm{LSE}} $'s bottlenecks.

Two techniques of $ \mathrm{IFC}_{\mathrm{LSE}} $ limit its search speed: voting and Hamming embedding. Voting requires large MA numbers ($ S $ and $ W $) to ensure that more positive results are returned and their voting scores are the highest, but it increases the number of images in the candidate set. Hamming embedding generates multiple binary codes based on assigned visual words, which calculates distances for most images more than once. We observe that the majority of the search time in $ \mathrm{IFC}_{\mathrm{LSE}} $ is spent on calculating the Hamming distances of embedding codes, which is the major bottleneck for efficiency. Our focus is on reducing the number of these calculations.

Our solution is to calculate a unique binary code for each image, to replace the embedding codes and to rank the returned images directly by Hamming distances. The benefits are two-fold. First, there is a need to ensure that only positive results can be found in the candidate set. Hence, maintaining the highest frequency of positive results becomes unnecessary, which can lower the MA numbers and reduce the amount of images in the candidate set. Second, only one distance needs to be calculated for each image. Both aspects lead to lower computational costs.

\begin{figure}[t]
\begin{center}
   \includegraphics[width=0.84\linewidth]{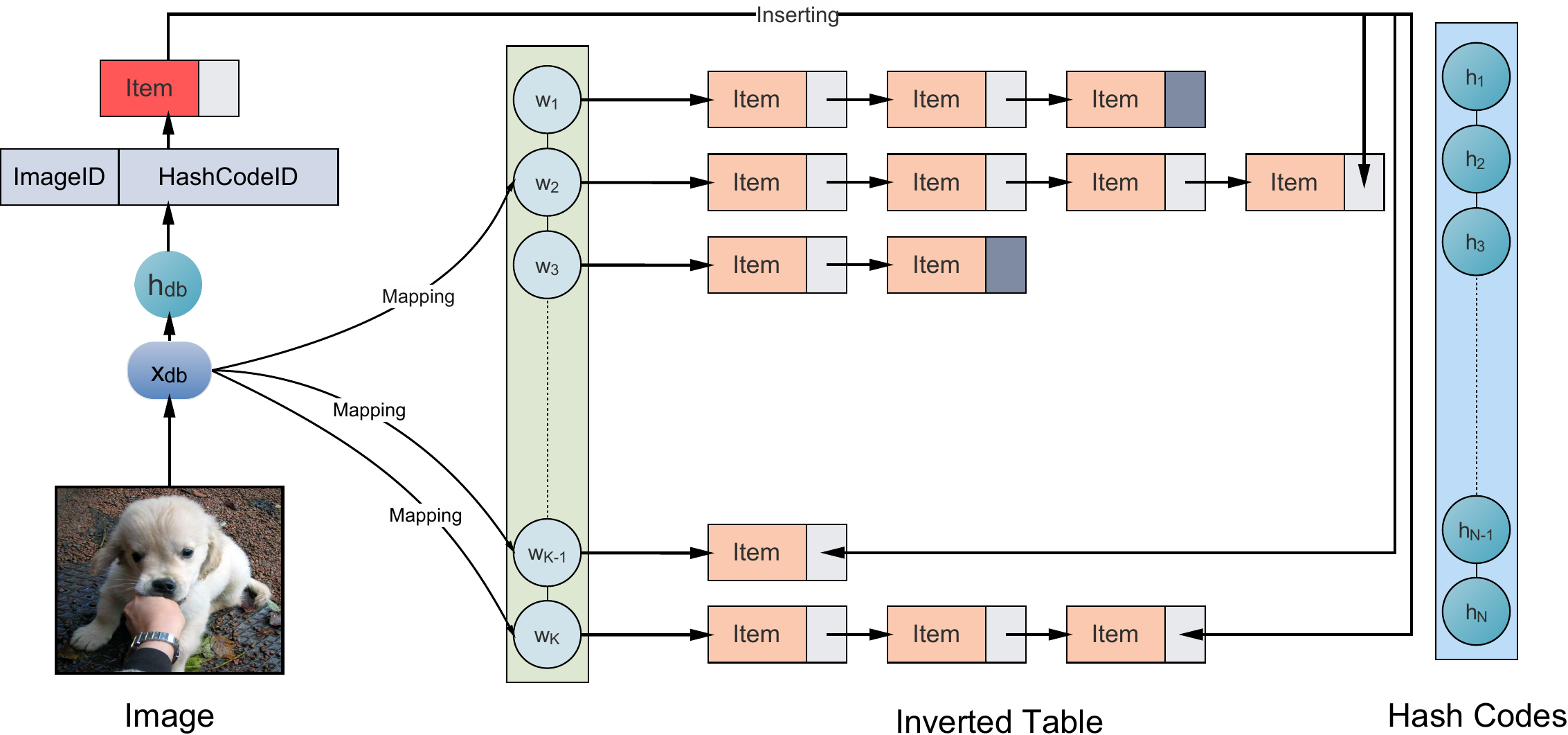}
\end{center}
   \caption{Illustration of $ \mathrm{IFC}_{\mathrm{HASH}} $ indexing structure. We calculate a hash code $ h_{db} $ for each feature vector $ x_{db} $ and store the hash codes individually. Each item of the inverted table consists the IDs of the image and its hash code.}
\label{fig:IFCHASH}
\end{figure}

The indexing structure of this improved scheme is illustrated in Fig. \ref{fig:IFCHASH}, denoted as $ \mathrm{IFC}_{\mathrm{HASH}} $. In addition, some minor changes have been made to its intrinsic structure. We peel off the hash codes from the inverted table and store them individually to save memory. In each item, we only keep the IDs of the image and its hash code. The complete procedures of $ \mathrm{IFC}_{\mathrm{HASH}} $ are summarized as follows:

\begin{enumerate}[1.]
\setlength{\itemindent}{1em}
\setlength{\itemsep}{1ex}
\item Construct a large visual dictionary by performing PKM clustering on CNN vectors.
\item Generate hash codes for the database images.
\item Assign each database image to its $ S $ nearest visual words. For each image, make an item containing the IDs of the image and its hash code. Then, insert the item into the lists of these visual words.
\item Each query looks into the lists of its $ W $ nearest visual words and the items inside are returned. The images of these items are returned and sorted by Hamming distances between the query's and their hash codes.
\end{enumerate}

\section{Experiments}
\label{sec:Experiments}                                                            

The experimental results and analysis are presented in this section. To validate the efficiency of the proposed framework, we perform experiments on four public benchmarks including three unsupervised and one supervised datasets. The proposed schemes are first tasked to small-scale image retrieval to evaluate the influence of the main parameters. Then, a large-scale retrieval task is considered by mixing a distracted dataset into the four benchmarks.

\subsection{Benchmarks}
\label{subsec:Benchmarks}

We use four small-scale datasets, \emph{i.e.}, Holiday, Oxford and UKbench as the unsupervised benchmarks, and NUS-WIDE as a supervised benchmark. MIRFlickr, a large-scale dataset, is used as the distractor. Details of the datasets are described as follows:

\textbf{Holiday} \cite{jegou2008hamming} is an image dataset containing personal holiday photos. It contains 1,491 images in total, separated into 500 groups. Each group represents a specific scene or object. The first image of each group is the query image, all others are the correct retrieval results.

\textbf{Oxford} \cite{philbin2007object} is a building dataset containing 5,062 images downloaded from Flickr by searching for particular Oxford landmarks. The dataset is manually annotated to generate a comprehensive ground truth for 11 different landmarks, each with 5 possible queries. Hence, there are 55 queries in total.

\textbf{UKbench} \cite{nister2006scalable} contains 10,200 images. It contains four-image groups, each represents the same object from different viewpoints. The first image in each group is the query image, which gives 2,550 queries in total.

\textbf{NUS-WIDE} \cite{NUSWIDE_gao2013visual} contains 269,648 images from Flickr, each of which is associated with at least one of 81 tags. The most frequent 21 tags are considered, resulting in 89,528 images. It is partitioned into two subsets, with 5\% as the query set and the remaining as the database set. Two images are a true match if they share at least one common tag.

\textbf{MIRFlickr} \cite{huiskes2008mir} contains 1,000,000 images downloaded from Flickr. We use them in our experiments as distractors to test the performance of large-scale image retrieval.

\subsection{Experimental Setup}
\label{subsec:ExperimentalSetup}

\textbf{Feature:} Several CNN features are extracted to test the adaptability of the proposed indexing framework, which includes AlexNet \cite{cnn_krizhevsky2012imagenet}, VGG16, VGG19 \cite{VGG_simonyan2014very}, GoogLeNet \cite{GoogLeNet_szegedy2015going}, ResNet50, ResNet101 and ResNet152 \cite{he2015deep}. The former three features are 4096D and the latter four ones are 1000D. All the features are extracted by Caffe \cite{jia2014caffe} using the pre-trained models from the ImageNet \cite{ImageNet_deng2009imagenet} database. \\

\noindent\textbf{Metric:} We use Mean Average Precision (MAP) as the accuracy metric. Given a query and a set of $ R $ retrieved results, the value of \textit{Average Precision} (AP) is defined as:
\begin{align}
\label{equation:ap}
\text{AP} = \frac{1}{l}\sum\nolimits_{r=1}^{R}P(r)\iota(r)
\end{align}
where $ l $ is the number of positive results in the retrieved set. $ P(r) $ denotes the precision of the top $ r $ retrieved documents, and $ \iota(r)=1 $ if the $ r^{\text{th}} $ retrieved document is a true positive, but $ \iota(r)=0 $ otherwise. The AP values over all the queries are averaged in the query set to obtain the MAP score. The larger the MAP, the better the accuracy. Note that the query image is ignored in both the retrieval results and the ground-truth to make MAP value more reasonable.

The search time in the on-line phase is used as the speed metric. It includes the running time to perform the entire search, including assignment, voting and ranking. A shorter time indicates better efficiency, and is calculated by averaging the results of all the queries.\\

\noindent\textbf{Platform:} All experiments are performed on a server with an Intel Xeon 2.5GHz CPU, 32GB RAM and Windows Server 2008 R2 x64 operating system.

\subsection{Compared Schemes}
\label{subsec:ComparedScheme}

Six schemes (including the proposed schemes) in total are compared in the experiments, which include:

\begin{itemize}
\item \textbf{\textit{Brute Force}}: Brute-force search.
\item $ \mathrm{\textit{\textbf{IFC}}}_{\mathrm{\textit{\textbf{LSE}}}} $ (proposed)
\item \textbf{\textit{LSH}}: Locality Sensitive Hashing~\cite{hashing_datar2004locality}.
\item $ \mathrm{\textit{\textbf{IFC}}}_{\mathrm{\textit{\textbf{LSH}}}} $: $ \mathrm{IFC}_{\mathrm{HASH}} $ whose codes are LSH (proposed).
\item \textbf{\textit{VDSH}}: Very Deep Supervised Hashing~\cite{VDSH_zhang2016efficient}.
\item $ \mathrm{\textit{\textbf{IFC}}}_{\mathrm{\textit{\textbf{VDSH}}}} $: $ \mathrm{IFC}_{\mathrm{HASH}} $ whose codes are VDSH (proposed).
\end{itemize}

The code for VDSH is provided by the author of \cite{VDSH_zhang2016efficient}. The others are implemented by ourselves. In the following experiments, we first test the proposed schemes on small-scale image retrieval tasks to observe the effect of the key parameters. Then, we compare the results to other schemes on both small-scale and large-scale retrievals to validate their effectiveness and efficiency.

\subsection{Effect of Key Parameters}
\label{subsec:EffectOfKeyParameters}

There are four parameters that mainly affect the precision and speed of the proposed schemes -- the MA numbers $ S $ and $ W $, the embedding code length $ L $, and the Hamming distance threshold $ T $. To test the sensitivity of the proposed schemes to these parameters, we perform experiments on three benchmarks: Holiday, Oxford and UKbench and take $ \mathrm{IFC}_{\mathrm{LSH}} $ as an example of $ \mathrm{IFC}_{\mathrm{HASH}} $.

\begin{figure}[t]
\begin{center}
\subfigure[]{\includegraphics[width=0.31\linewidth]{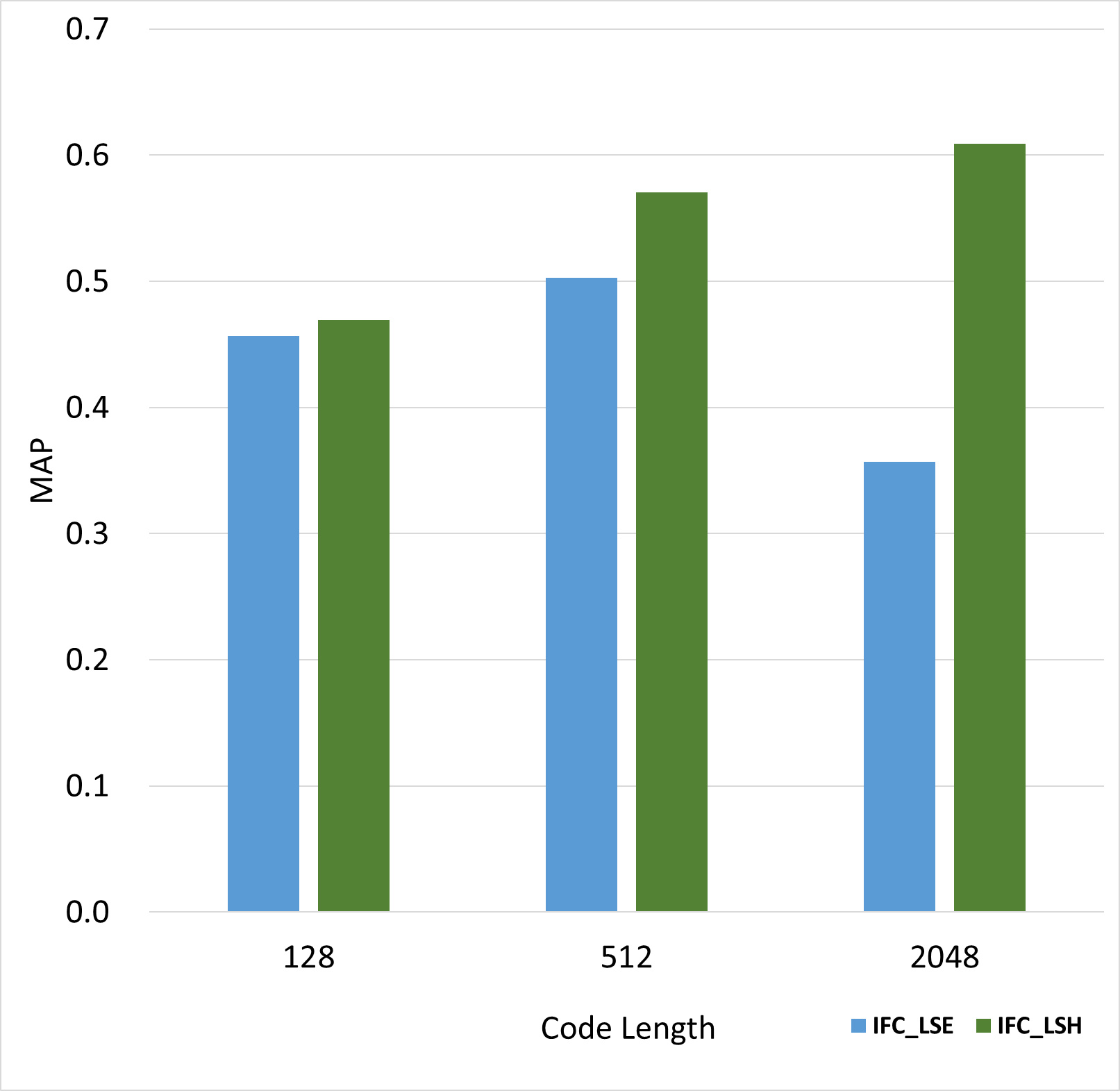}}
\subfigure[]{\includegraphics[width=0.31\linewidth]{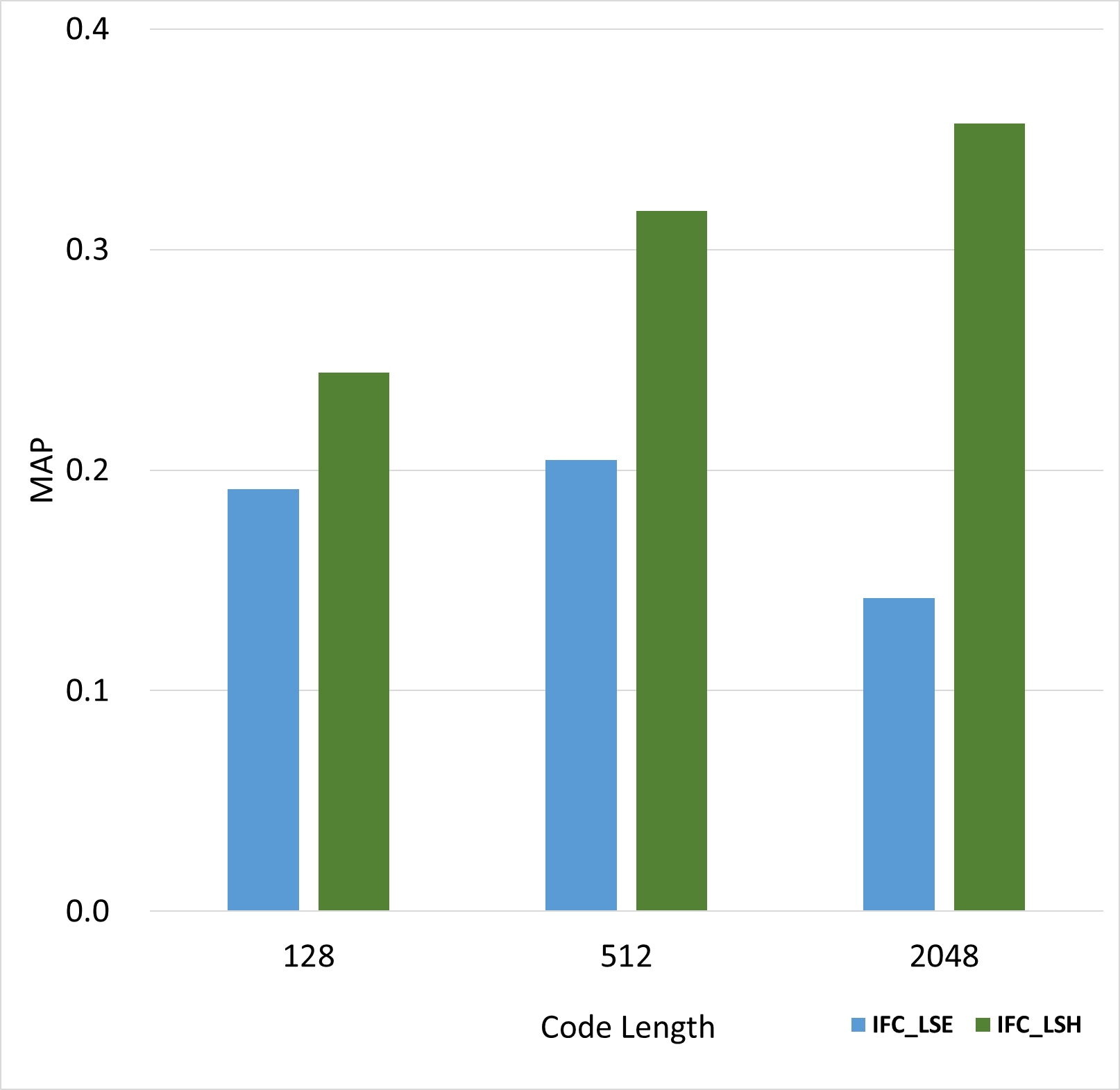}}
\subfigure[]{\includegraphics[width=0.31\linewidth]{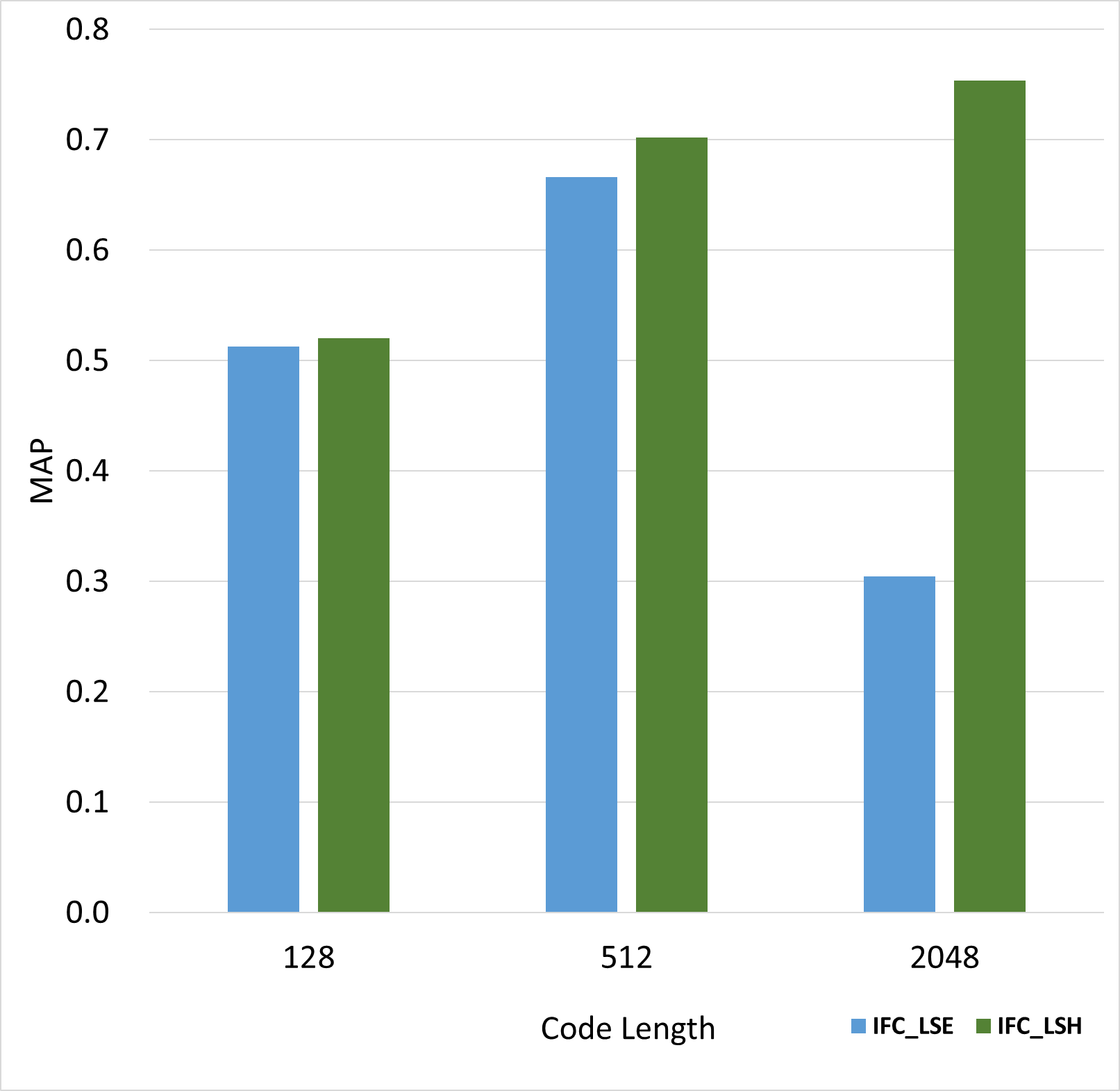}}
\end{center}
	\caption{Illustrations of MAP value variation with code length on (a) Holiday, (b) Oxford and (c) UKbench datasets.}
\label{fig:CodeLength}
\end{figure}

The sensitivity to code length $ L $ of the proposed schemes on the three benchmarks is illustrated in Fig. \ref{fig:CodeLength}, where other parameters are set proper values according to the conclusions drawn later. Clearly, for $ \mathrm{IFC}_{\mathrm{LSE}} $, a medium length of embedding code achieves the best performance. It is reasonable to reach this conclusion since a median-length embedding code best balances missed and false matches. While for $ \mathrm{IFC}_{\mathrm{LSH}} $, the best performance is achieved at the longest code. It fits the conclusion of~\cite{hashing_datar2004locality} that a longer code improves the probability of mapping similar vectors into a same code. But a longer code also brings higher computation costs. To balance precision and speed, $ L $ is set at $ 512 $ for both $ \mathrm{IFC}_{\mathrm{LSE}} $ and $ \mathrm{IFC}_{\mathrm{LSH}} $ in subsequent experiments.

\begin{figure}[t]
\begin{center}
\subfigure[$ \mathrm{IFC}_{\mathrm{LSE}} $]{\includegraphics[width=0.45\linewidth]{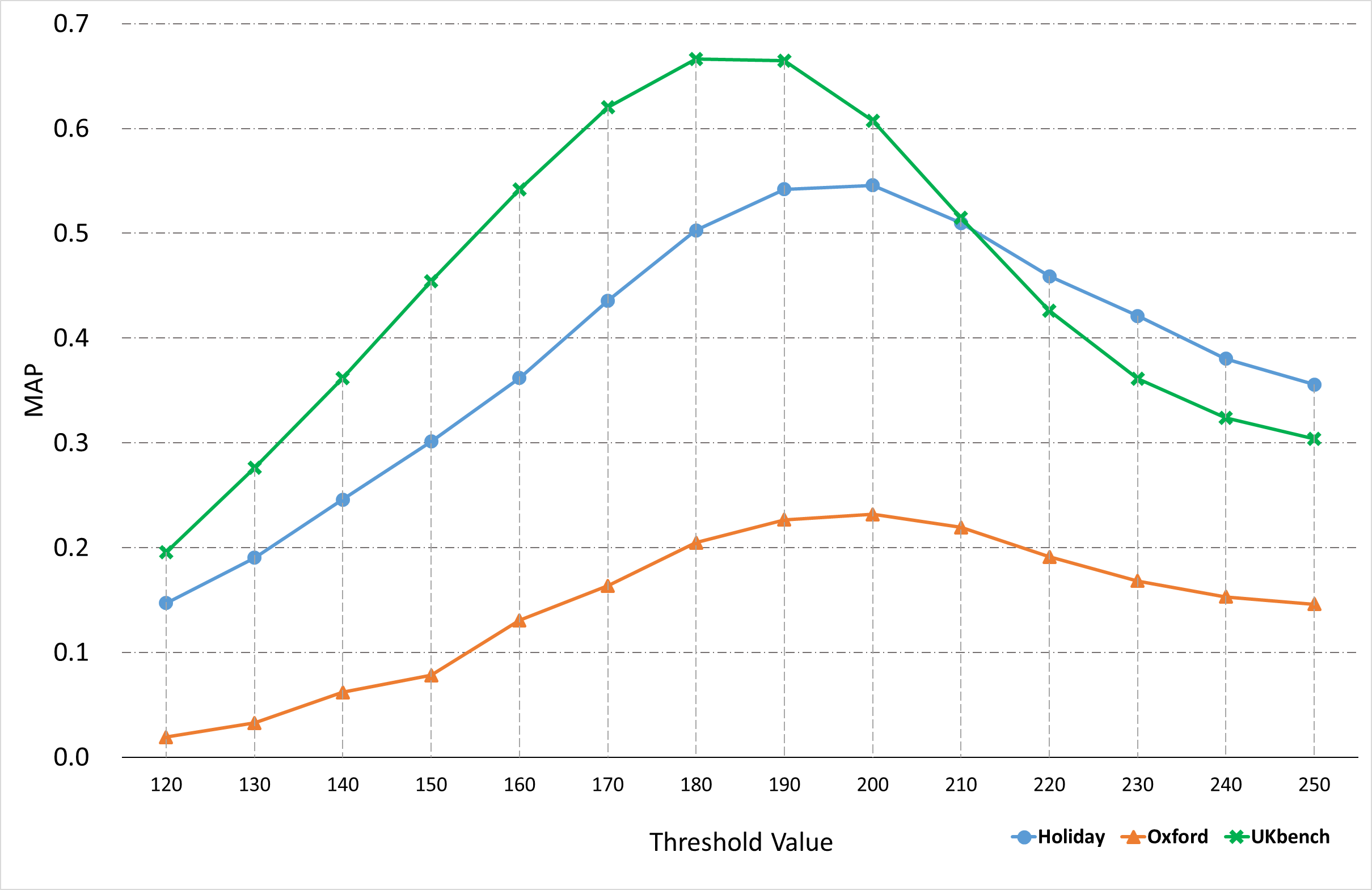}}
\subfigure[$ \mathrm{IFC}_{\mathrm{LSH}} $]{\includegraphics[width=0.45\linewidth]{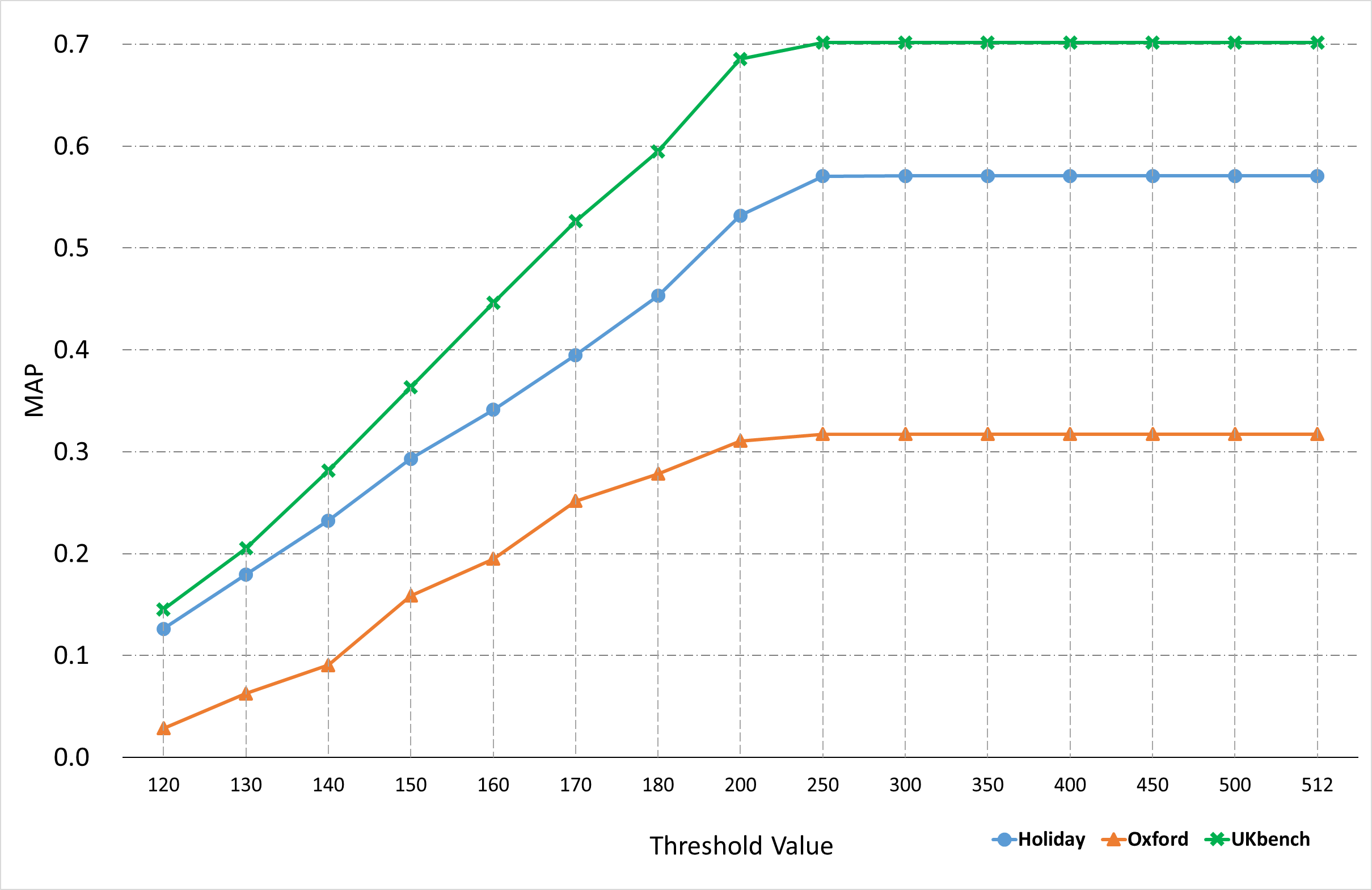}}
\end{center}
	\caption{Illustration of MAP value variation with Hamming distance threshold for (a) $ \mathrm{IFC}_{\mathrm{LSE}} $, (b) $ \mathrm{IFC}_{\mathrm{LSH}} $.}
\label{fig:Threshold}
\end{figure} 

To test sensitivity to the Hamming distance threshold $ T $, we also carry out a series of experiments. The results on $ \mathrm{IFC}_{\mathrm{LSE}} $ are illustrated in Fig. \ref{fig:Threshold}(a) and we observe that $ T $ is properly set at $ 200, 200, 180 $ for Holiday, Oxford and UKbench separately. When $ T $ is small, many correct results are rejected by mistake. By contrast, when $ T $ is too large, too many wrong results are reserved. Empirically, the best threshold is determined to sit at about $ 30 \% $ of the code length. While for $ \mathrm{IFC}_{\mathrm{LSH}} $, whose results are illustrated in Fig. \ref{fig:Threshold}(b), precision is improved with $ T $ until it reaches a specific value. Unlike $ \mathrm{IFC}_{\mathrm{LSE}} $, $ \mathrm{IFC}_{\mathrm{LSH}} $ generates a unique code for each image and ranks the returned images directly according to the Hamming distance. This threshold simply removes some false matches, and the proper values are $ 300, 250, 250 $ for the three benchmarks. This means that distances between correct results and queries are below these thresholds.

\begin{figure}[t]
\begin{center}
\subfigure[]{\includegraphics[width=0.3\linewidth]{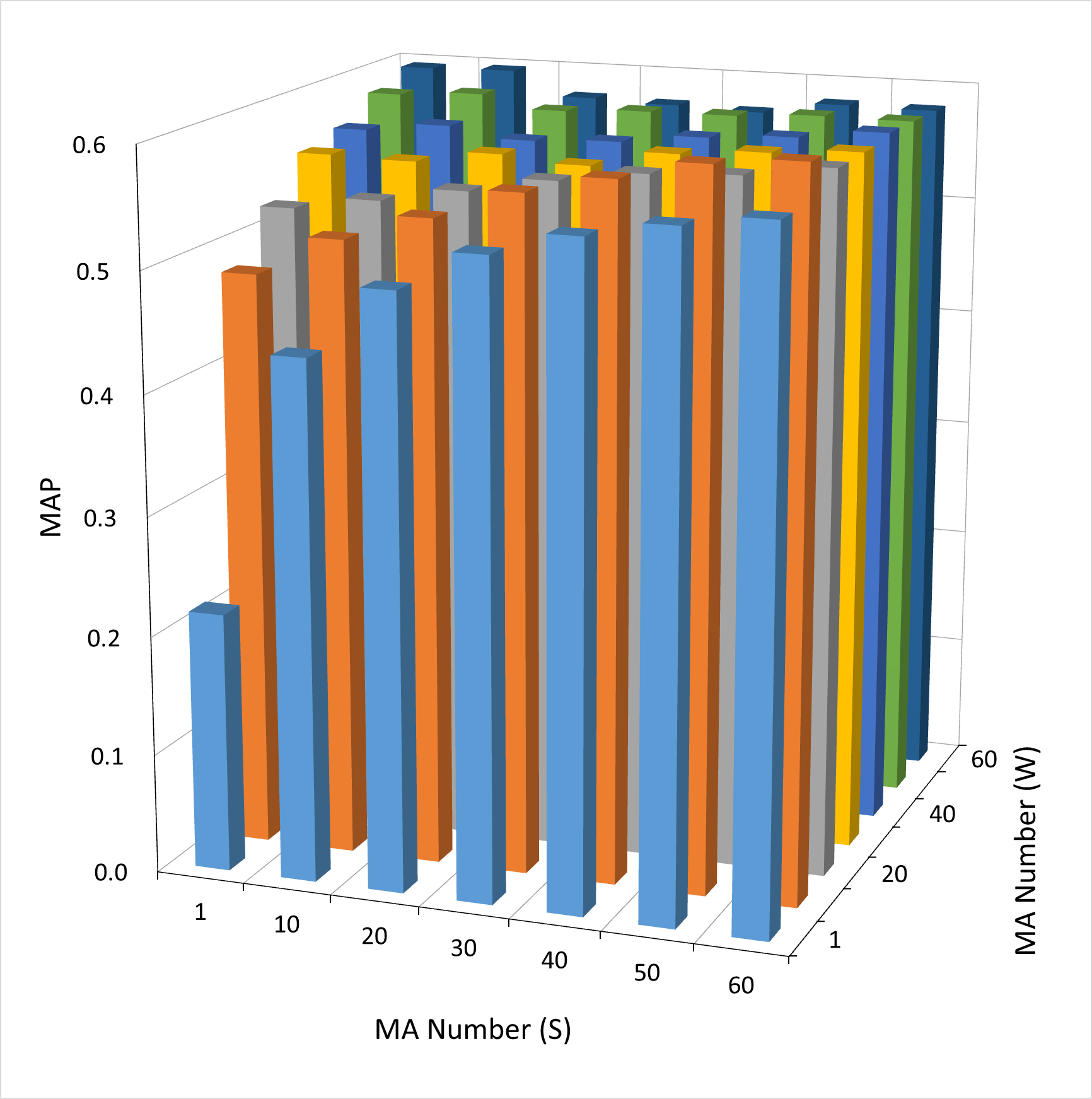}}
\subfigure[]{\includegraphics[width=0.3\linewidth]{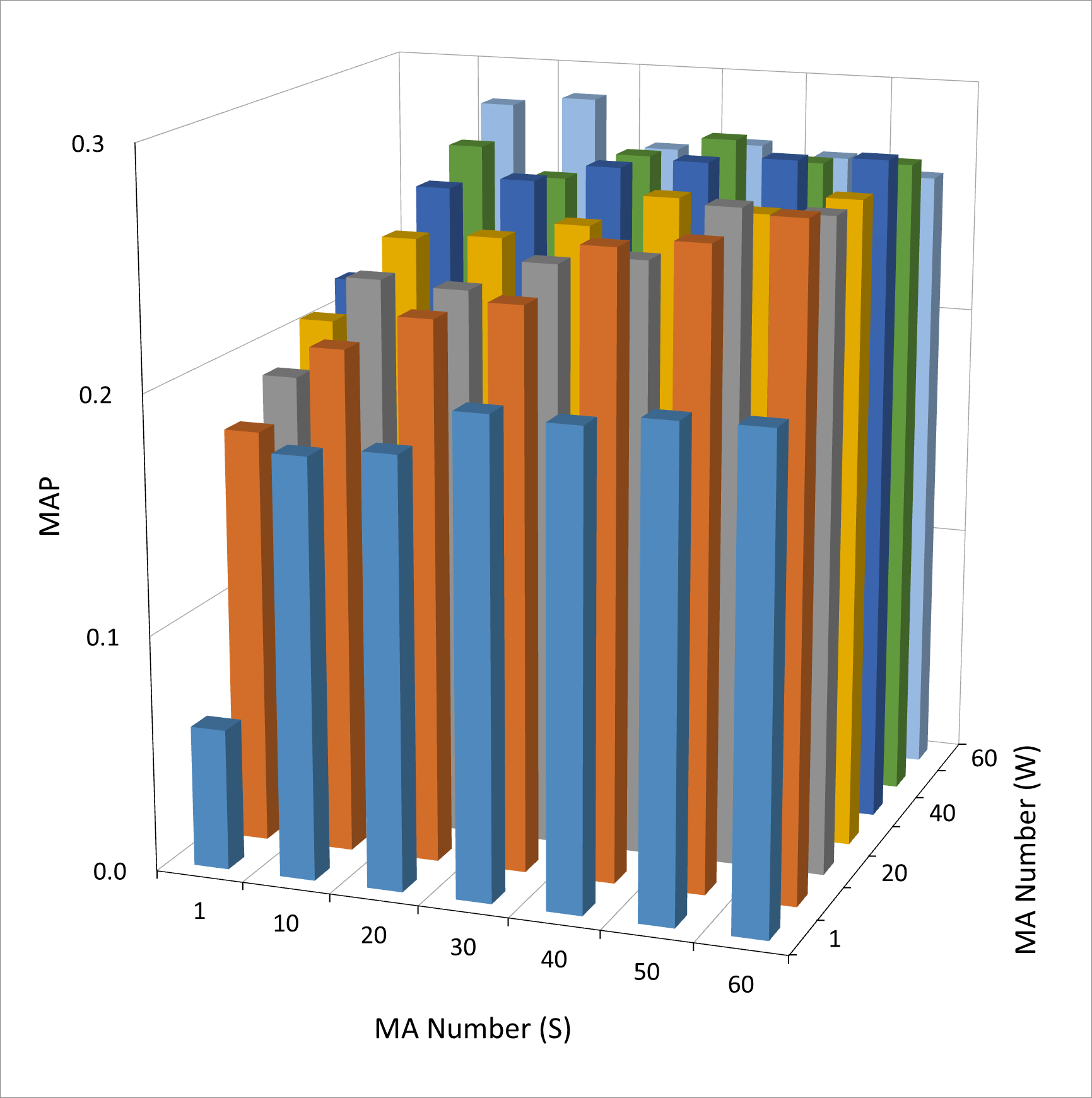}}
\subfigure[]{\includegraphics[width=0.3\linewidth]{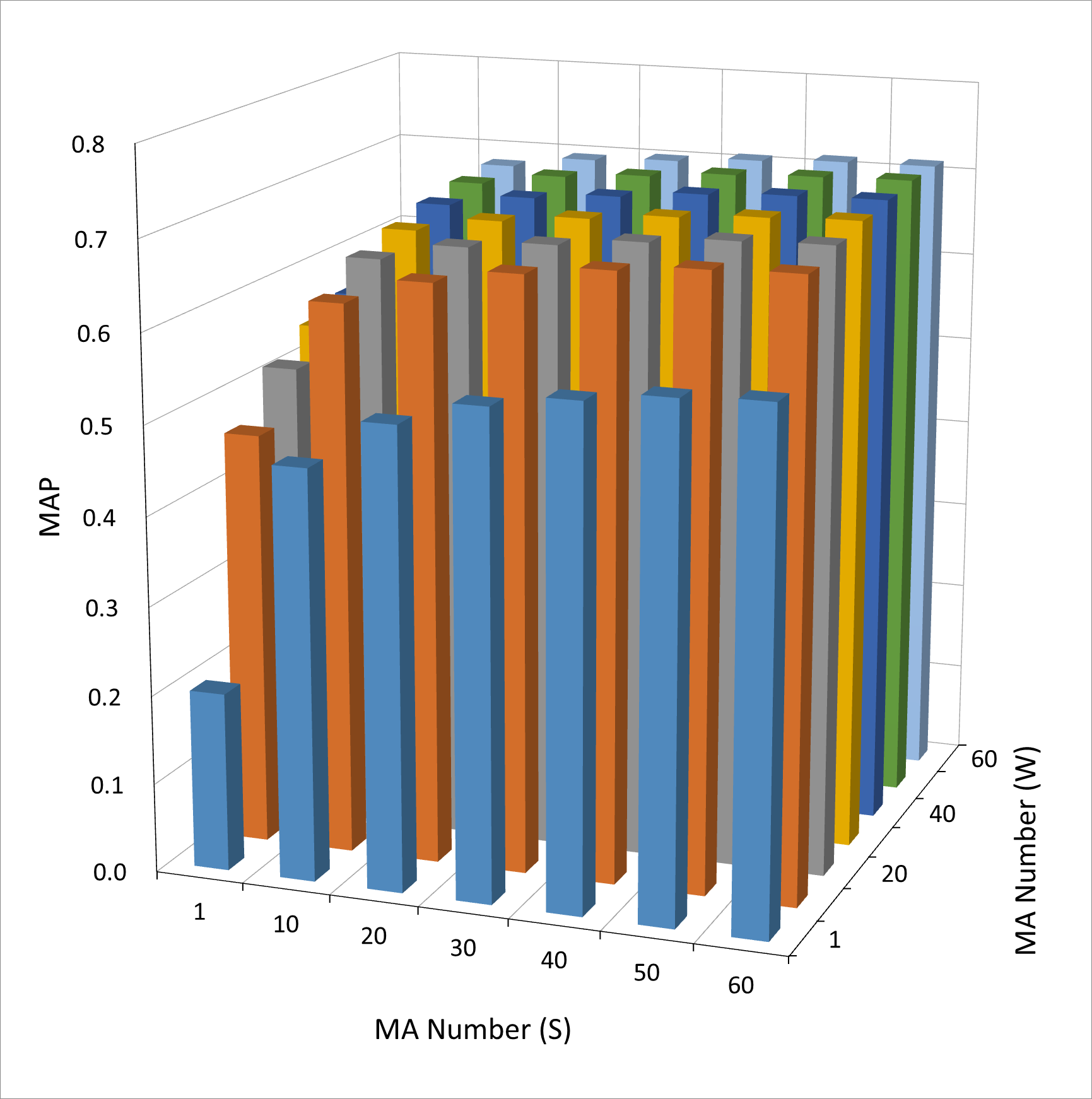}}
\end{center}
	\caption{Illustrations of MAP value variation with MA numbers ($ S $ and $ W $) for $ \mathrm{IFC}_{\mathrm{LSE}} $ on (a) Holiday, (b) Oxford and (c) UKbench datasets.}
\label{fig:IFCLSESWInteraction}
\end{figure}

\begin{figure}[t]
\begin{center}
\subfigure[]{\includegraphics[width=0.3\linewidth]{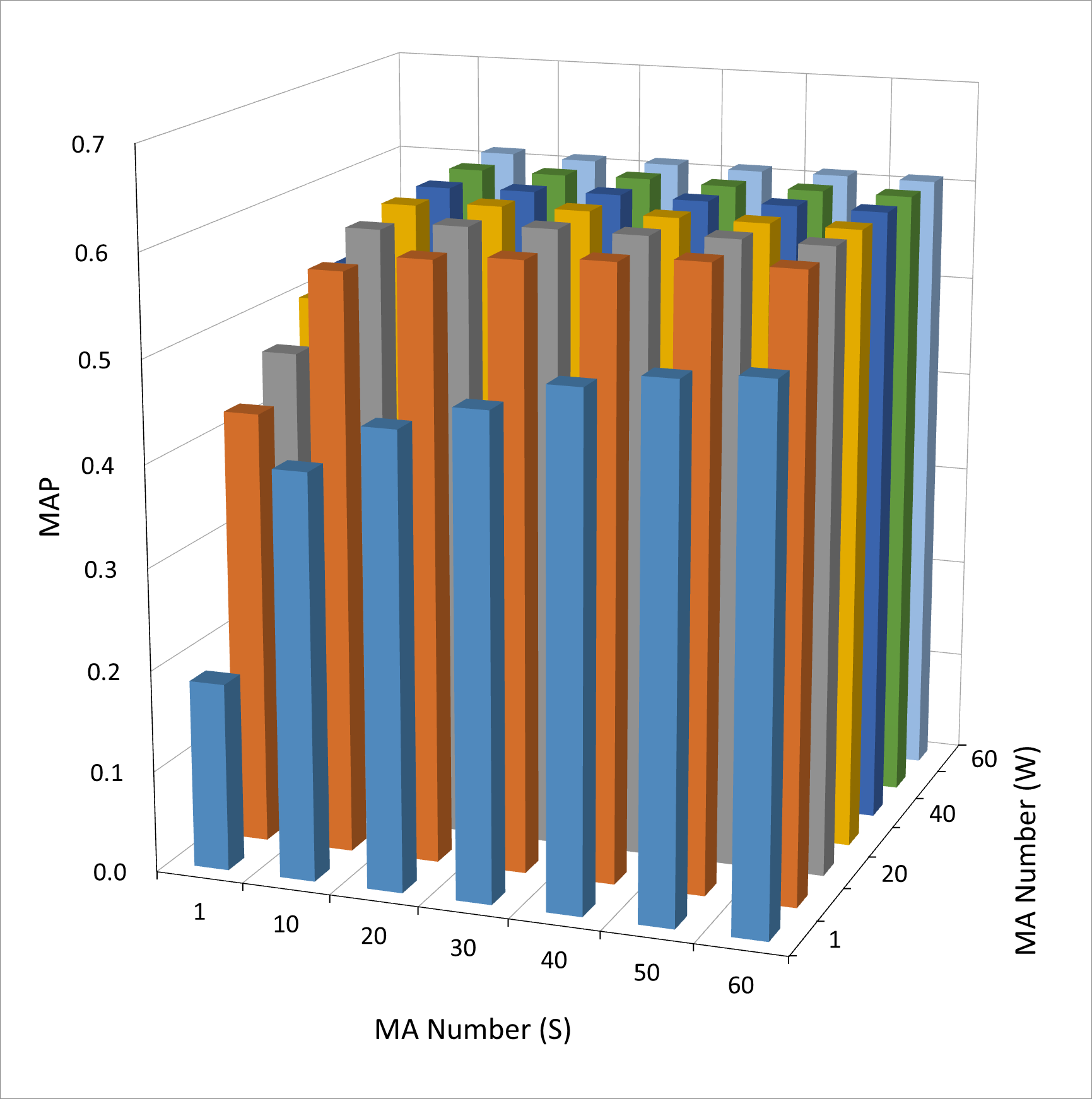}}
\subfigure[]{\includegraphics[width=0.3\linewidth]{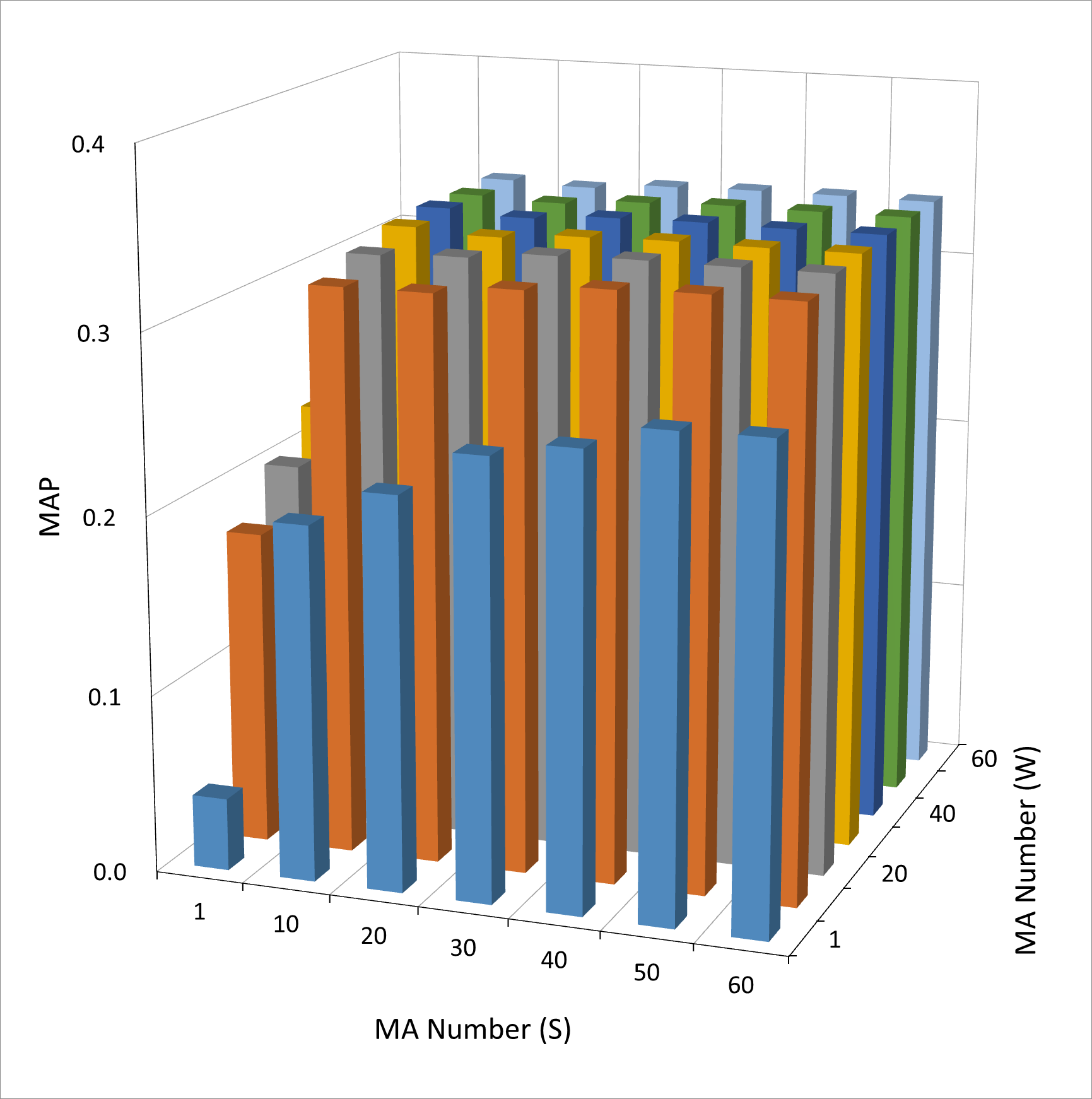}}
\subfigure[]{\includegraphics[width=0.3\linewidth]{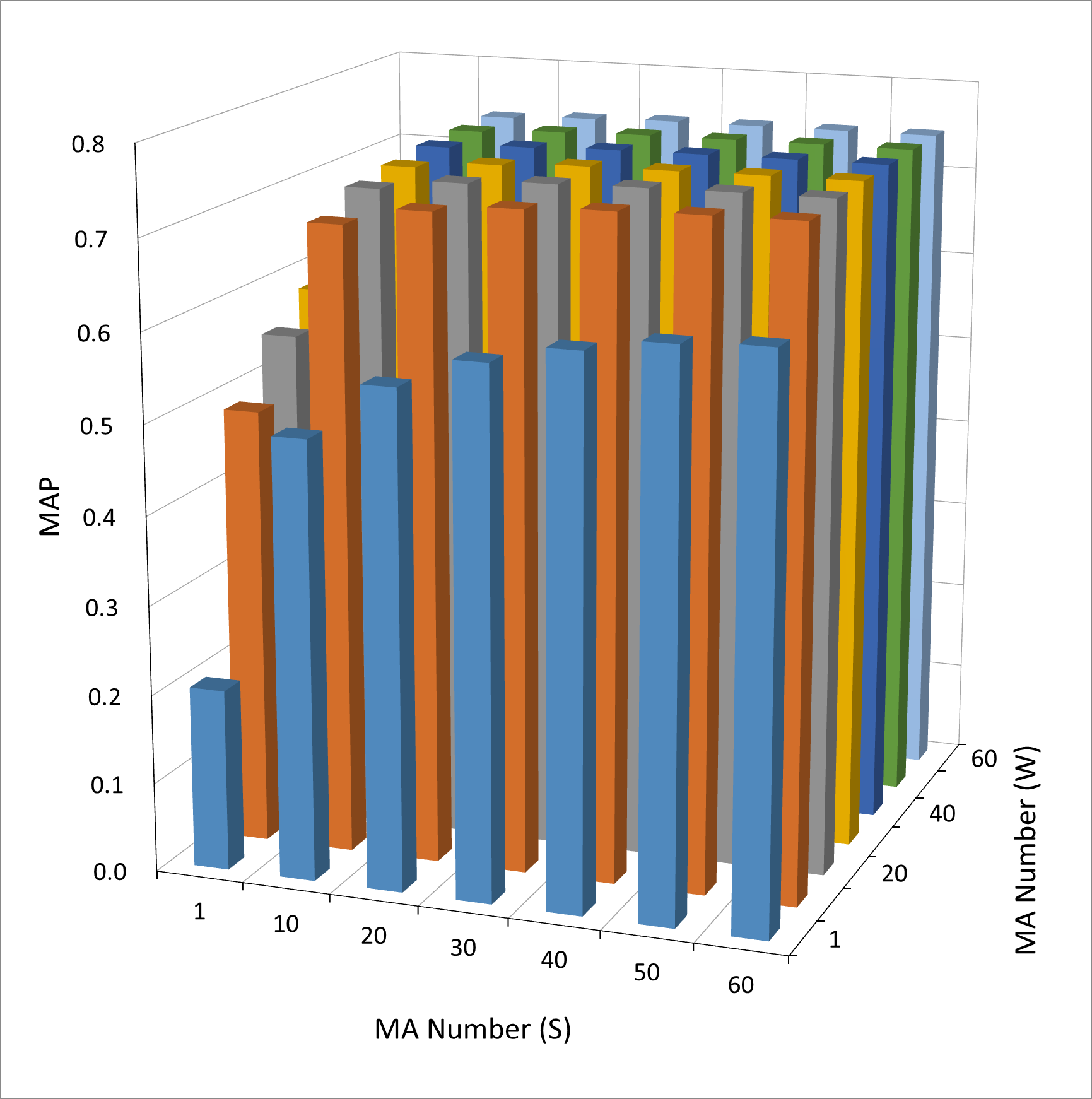}}
\end{center}
	\caption{Illustrations of MAP value variation with MA numbers ($ S $ and $ W $) for $ \mathrm{IFC}_{\mathrm{LSH}} $ on (a) Holiday, (b) Oxford and (c) UKbench datasets.}
\label{fig:IFCLSHSWInteraction}
\end{figure}

Both MA numbers, \emph{i.e.}, $ S $ and $ W $, indicate the number of candidates that take part in the final ranking also influence the precision and speed of the proposed schemes. The sensitivity to $ S $ and $ W $ of $ \mathrm{IFC}_{\mathrm{LSE}} $ and $ \mathrm{IFC}_{\mathrm{LSH}} $ are illustrated in Figs. \ref{fig:IFCLSESWInteraction} and \ref{fig:IFCLSHSWInteraction} respectively, where experiments are performed under different combinations of the two parameters. We observe that, for both schemes, higher precision is achieved when either of the parameters is equal to a large value. This satisfies the description in Section \ref{sec:detail-of-proposed-framework}, and in our experiments we simply set $ S $ and $ W $ as equal when choosing proper values.

\begin{figure}[t]
\begin{center}
\subfigure[]{\includegraphics[width=0.45\linewidth]{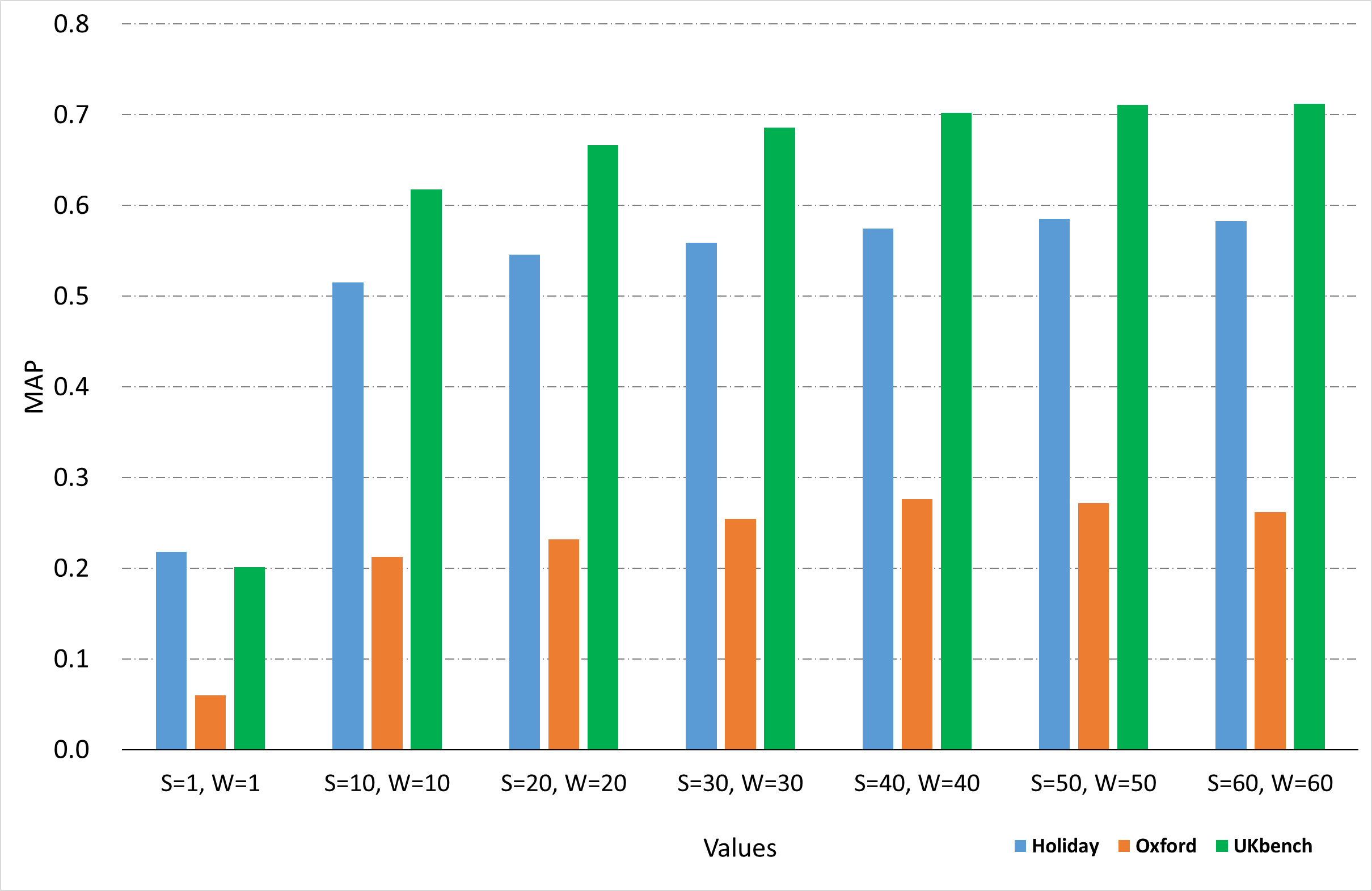}}
\subfigure[]{\includegraphics[width=0.45\linewidth]{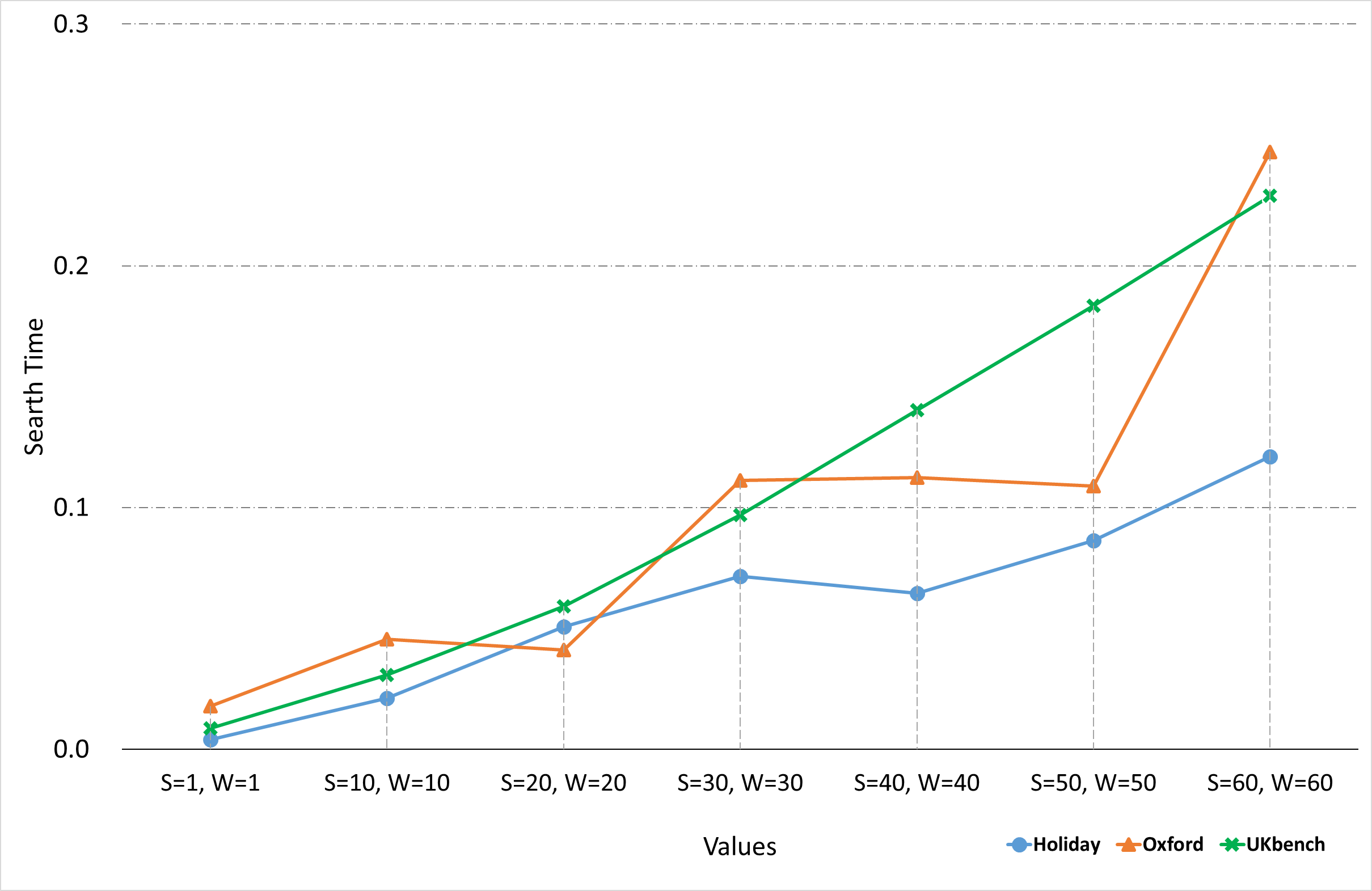}}
\end{center}
	\caption{Illustrations of (a) MAP value and (b) search time variation with MA numbers ($ S $ and $ W $) for $ \mathrm{IFC}_{\mathrm{LSE}} $, when they are equal, on Holiday, Oxford and UKbench datasets.}
\label{fig:IFCLSESWSensitivity}
\end{figure}

\begin{figure}[!ht]
\begin{center}
\subfigure[]{\includegraphics[width=0.45\linewidth]{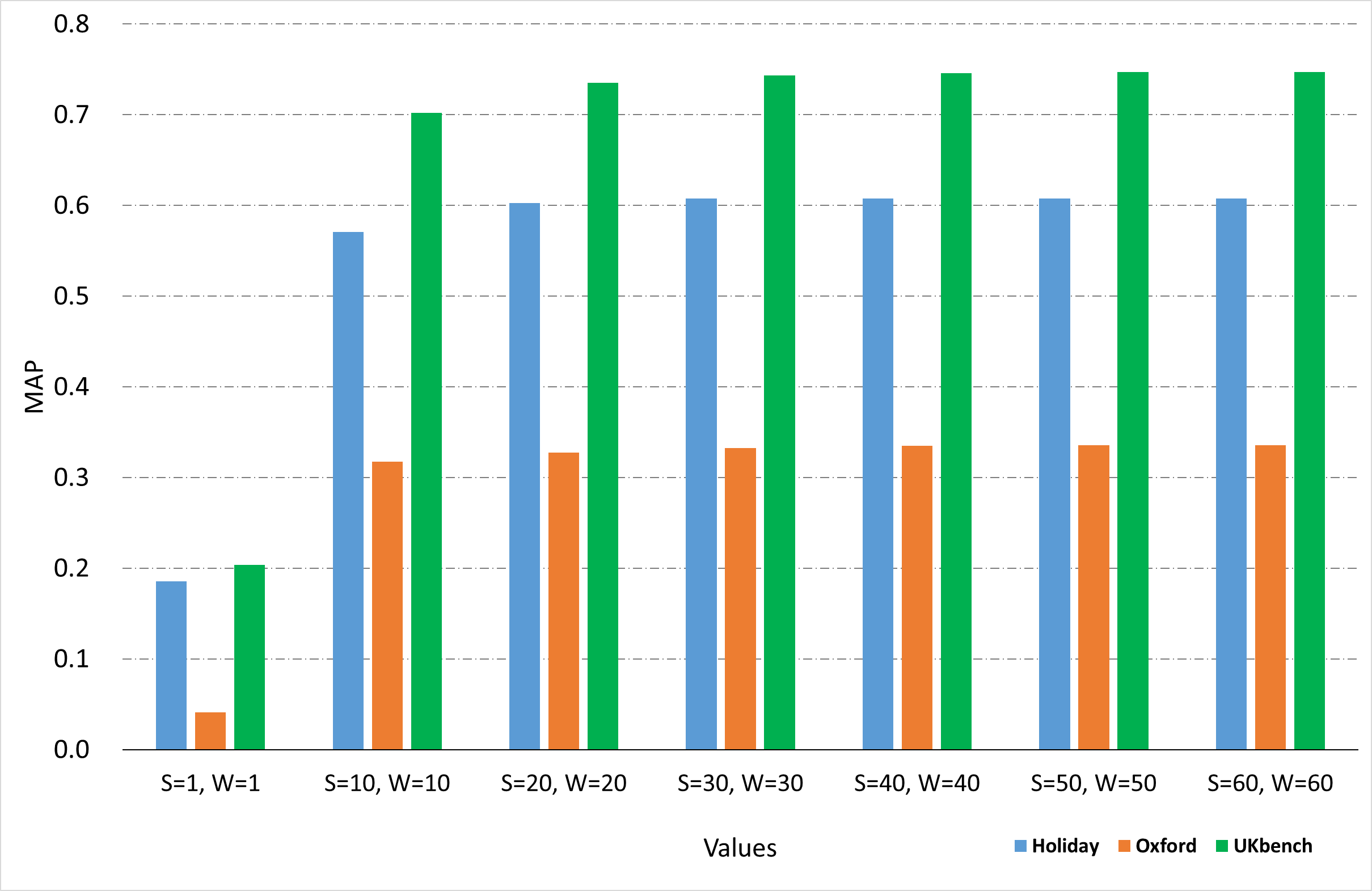}}
\subfigure[]{\includegraphics[width=0.45\linewidth]{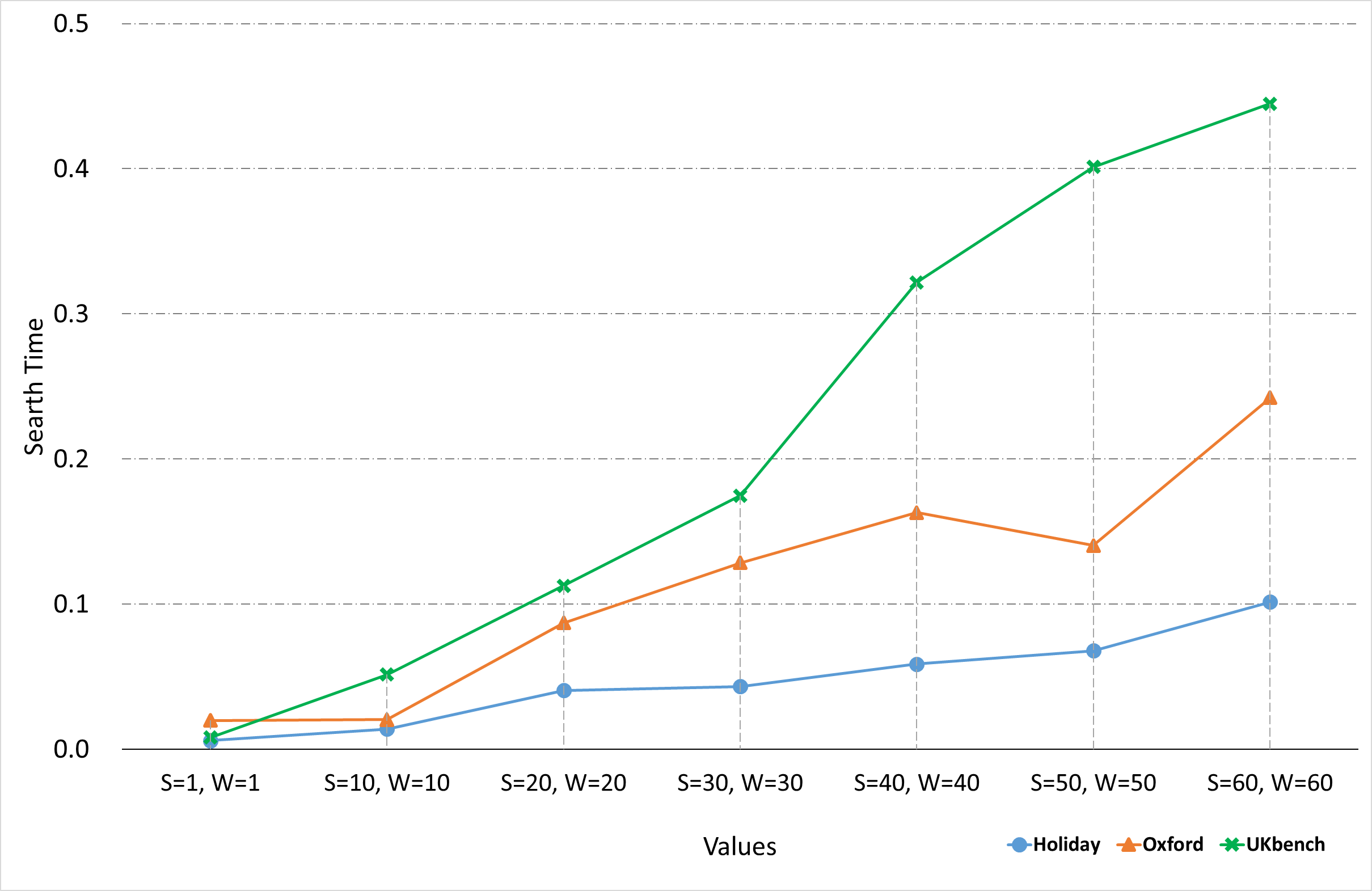}}
\end{center}
	\caption{Illustrations of (a) MAP value and (b) search time variation with MA numbers ($ S $ and $ W $) for $ \mathrm{IFC}_{\mathrm{LSH}} $, when they are equal, on Holiday, Oxford and UKbench datasets.}
\label{fig:IFCLSHSWSensitivity}
\end{figure}

The experimental results of sensitivity to $ S $ and $ W $, when they are equal, are illustrated in Figs. \ref{fig:IFCLSESWSensitivity} and \ref{fig:IFCLSHSWSensitivity} for the two proposed schemes respectively. They intuitively reflect that the precision of both the schemes is improved with an increase of the two parameters, but they also increases the search time. However, when they reach a specific value, improvements to precision are not obvious. We observe that $ 20 $ is a proper value for $ S $ and $ W $ in $ \mathrm{IFC}_{\mathrm{LSE}} $ to balance its accuracy and search time. While for $ \mathrm{IFC}_{\mathrm{LSH}} $, $ 10 $ is sufficient to achieve good precision and speed.

\subsection{Comparison on Unsupervised Benchmarks}

In this subsection, the proposed schemes are compared to several methods on the three unsupervised benchmarks to test effectiveness and efficiency. We also perform experiments on multiple CNN features to validate the adaptability of the proposed schemes. $ \mathrm{IFC}_{\mathrm{LSH}} $ is taken as an example of $ \mathrm{IFC}_{\mathrm{HASH}} $ since LSH is unsupervised and it is another baseline. For both proposed schemes, the parameters are set according to the conclusions drawn in Section \ref{subsec:EffectOfKeyParameters}.

\begin{table}[t]
\begin{center}
\caption{Evaluation on three unsupervised benchmarks for small-scale retrieval}
\label{table:unsupervised_small}
\begin{tabular}{|c|c|>{\columncolor{gray!25}}c|c|>{\columncolor{gray!25}}c|c|>{\columncolor{gray!25}}c|}
\hline
\multirow{3}{*}{Method}
&\multicolumn{6}{c|}{Dataset} \\
\cline{2-7}
&\multicolumn{2}{c|}{Holiday} &\multicolumn{2}{c|}{Oxford} &\multicolumn{2}{c|}{UKbench} \\
\hhline{~------}
& MAP & Time & MAP & Time & MAP & Time \\
\hline\hline
\multicolumn{7}{|c|}{AlexNet} \\
\hline
Brute Force & 0.5728 & 0.012 & 0.3185 & 0.132 & 0.7221 & 0.240 \\
\hline
LSH & 0.6027 & 0.012 & 0.3358 & 0.034 & 0.7461 & 0.049 \\
\hline
$ \mathrm{IFC}_{\mathrm{LSE}} $ & 0.5459 & 0.051 & 0.2318 & 0.041 & 0.6662 & 0.052 \\
\hline
$ \mathrm{IFC}_{\mathrm{LSH}} $ & 0.5707 & 0.009 & 0.3174 & 0.010 & 0.7019 & 0.028 \\
\hline\hline
\multicolumn{7}{|c|}{VGG16} \\
\hline
Brute Force & 0.6593 & 0.013 & 0.2624 & 0.122 & 0.8236 & 0.231 \\
\hline
LSH & 0.6737 & 0.011 & 0.2423 & 0.030 & 0.8021 & 0.052 \\
\hline
$ \mathrm{IFC}_{\mathrm{LSE}} $ & 0.5352 & 0.040 & 0.1364 & 0.092 & 0.6439 & 0.061 \\
\hline
$ \mathrm{IFC}_{\mathrm{LSH}} $ & 0.6093 & 0.023 & 0.2253 & 0.037 & 0.7610 & 0.026 \\
\hline\hline
\multicolumn{7}{|c|}{VGG19} \\
\hline
Brute Force & 0.6433 & 0.010 & 0.2206 & 0.126 & 0.8167 & 0.240 \\
\hline
LSH & 0.6659 & 0.034 & 0.2048 & 0.052 & 0.8002 & 0.053 \\
\hline
$ \mathrm{IFC}_{\mathrm{LSE}} $ & 0.5150 & 0.042 & 0.1643 & 0.078 & 0.6310 & 0.060 \\
\hline
$ \mathrm{IFC}_{\mathrm{LSH}} $ & 0.5856 & 0.019 & 0.1782 & 0.045 & 0.7533 & 0.028 \\
\hline\hline
\multicolumn{7}{|c|}{GoogLeNet} \\
\hline
Brute Force & 0.6089 & 0.009 & 0.2173 & 0.109 & 0.7986 & 0.217 \\
\hline
LSH & 0.5847 & 0.074 & 0.2024 & 0.076 & 0.7488 & 0.049 \\
\hline
$ \mathrm{IFC}_{\mathrm{LSH}} $ & 0.5066 & 0.017 & 0.1789 & 0.027 & 0.6631 & 0.019 \\
\hline\hline
\multicolumn{7}{|c|}{ResNet50} \\
\hline
Brute Force & 0.6836 & 0.010 & 0.2677 & 0.105 & 0.8664 & 0.227 \\
\hline
LSH & 0.6675 & 0.010 & 0.2737 & 0.031 & 0.8375 & 0.050 \\
\hline
$ \mathrm{IFC}_{\mathrm{LSH}} $ & 0.5554 & 0.013 & 0.2571 & 0.042 & 0.7337 & 0.021 \\
\hline\hline
\multicolumn{7}{|c|}{ResNet101} \\
\hline
Brute Force & 0.6977 & 0.010 & 0.2814 & 0.106 & 0.8610 & 0.222 \\
\hline
LSH & 0.6729 & 0.031 & 0.2148 & 0.038 & 0.8422 & 0.051 \\
\hline
$ \mathrm{IFC}_{\mathrm{LSH}} $ & 0.5725 & 0.013 & 0.2070 & 0.039 & 0.7312 & 0.021 \\
\hline\hline
\multicolumn{7}{|c|}{ResNet152} \\
\hline
Brute Force & 0.6977 & 0.009 & 0.2837 & 0.106 & 0.8636 & 0.231 \\
\hline
LSH & 0.6860 & 0.008 & 0.2679 & 0.027 & 0.8366 & 0.050 \\
\hline
$ \mathrm{IFC}_{\mathrm{LSH}} $ & 0.5914 & 0.023 & 0.2566 & 0.031 & 0.7321 & 0.018 \\
\hline
\end{tabular}
\end{center}
\end{table}

\begin{table}[t]
\begin{center}
\caption{Evaluation on three unsupervised benchmarks for large-scale retrieval}
\label{table:unsupervised_large}
\begin{tabular}{|c|c|>{\columncolor{gray!25}}c|c|>{\columncolor{gray!25}}c|c|>{\columncolor{gray!25}}c|}
\hline
\multirow{3}{*}{Method}
&\multicolumn{6}{c|}{Dataset} \\
\cline{2-7}
&\multicolumn{2}{c|}{Holiday+MIRFlickr} &\multicolumn{2}{c|}{Oxford+MIRFlickr} &\multicolumn{2}{c|}{UKbench+MIRFlickr} \\
\hhline{~------}
& MAP & Time & MAP & Time & MAP & Time \\
\hline\hline
\multicolumn{7}{|c|}{AlexNet} \\
\hline
Brute Force & 0.4896 & 210.934 & 0.2557 & 241.002 & 0.6579 & 231.414 \\
\hline
LSH & 0.4884 & 3.324 & 0.2661 & 3.486 & 0.6975 & 3.449 \\
\hline
$ \mathrm{IFC}_{\mathrm{LSE}} $ & 0.3848 & 0.503 & 0.1759 & 0.241 & 0.6572 & 0.971 \\
\hline
$ \mathrm{IFC}_{\mathrm{LSH}} $ & 0.4747 & 0.203 & 0.2638 & 0.084 & 0.6667 & 0.299 \\
\hline\hline
\multicolumn{7}{|c|}{VGG16} \\
\hline
Brute Force & 0.4930 & 229.147 & 0.2079 & 258.215 & 0.7897 & 233.632 \\
\hline
LSH & 0.4474 & 3.285 & 0.1930 & 3.429 & 0.7477 & 3.314 \\
\hline
$ \mathrm{IFC}_{\mathrm{LSE}} $ & 0.1725 & 0.358 & 0.0797 & 0.528 & 0.5452 & 0.317 \\
\hline
$ \mathrm{IFC}_{\mathrm{LSH}} $ & 0.4249 & 0.122 & 0.1884 & 0.167 & 0.7172 & 0.109 \\
\hline\hline
\multicolumn{7}{|c|}{VGG19} \\
\hline
Brute Force & 0.4584 & 240.308 & 0.1511 & 257.909 & 0.7807 & 242.074 \\
\hline
LSH & 0.4235 & 3.424 & 0.1339 & 3.301 & 0.7430 & 3.314 \\
\hline
$ \mathrm{IFC}_{\mathrm{LSE}} $ & 0.1627 & 0.353 & 0.0928 & 0.460 & 0.5213 & 0.304 \\
\hline
$ \mathrm{IFC}_{\mathrm{LSH}} $ & 0.3991 & 0.126 & 0.1243 & 0.158 & 0.7080 & 0.084 \\
\hline\hline
\multicolumn{7}{|c|}{GoogLeNet} \\
\hline
Brute Force & 0.3522 & 246.233 & 0.1348 & 265.454 & 0.7192 & 239.007 \\
\hline
LSH & 0.2997 & 3.273 & 0.1293 & 3.315 & 0.6509 & 3.298 \\
\hline
$ \mathrm{IFC}_{\mathrm{LSH}} $ & 0.2859 & 0.085 & 0.1252 & 0.108 & 0.5946 & 0.057 \\
\hline\hline
\multicolumn{7}{|c|}{ResNet50} \\
\hline
Brute Force & 0.4725 & 248.600 & 0.2161 & 265.714 & 0.8324 & 247.769 \\
\hline
LSH & 0.4443 & 3.408 & 0.1972 & 3.411 & 0.7884 & 3.451 \\
\hline
$ \mathrm{IFC}_{\mathrm{LSH}} $ & 0.4034 & 0.089 & 0.1935 & 0.192 & 0.7034 & 0.068 \\
\hline\hline
\multicolumn{7}{|c|}{ResNet101} \\
\hline
Brute Force & 0.4814 & 245.295 & 0.2188 & 269.339 & 0.8294 & 242.765 \\
\hline
LSH & 0.4591 & 3.258 & 0.1522 & 3.422 & 0.7980 & 3.155 \\
\hline
$ \mathrm{IFC}_{\mathrm{LSH}} $ & 0.4211 & 0.086 & 0.1515 & 0.147 & 0.7044 & 0.048 \\
\hline\hline
\multicolumn{7}{|c|}{ResNet152} \\
\hline
Brute Force & 0.4888 & 248.398 & 0.2215 & 267.582 & 0.8333 & 250.913 \\
\hline
LSH & 0.4668 & 3.266 & 0.2091 & 3.422 & 0.7904 & 3.448 \\
\hline
$ \mathrm{IFC}_{\mathrm{LSH}} $ & 0.4351 & 0.082 & 0.2058 & 0.182 & 0.7029 & 0.062 \\
\hline
\end{tabular}
\end{center}
\end{table}

The experimental results are summarized in Tables \ref{table:unsupervised_small} and \ref{table:unsupervised_large}, from which we draw the following conclusions:

(1) LSH outperforms the other schemes and $ \mathrm{IFC}_{\mathrm{LSH}} $ comes close in terms of precision. We observe that LSH shows even better performance than brute force in most cases. This is mainly because the code length chosen in this case is large ($ L = 512 $), as explained in the last subsection. $ \mathrm{IFC}_{\mathrm{LSH}} $ is a little worse than LSH because, under the setting of MA numbers ($ S = 10, W = 10 $), some positive results are missing in the candidate set. 

(2) $ \mathrm{IFC}_{\mathrm{LSH}} $ and $ \mathrm{IFC}_{\mathrm{LSE}} $ succeed in improving search speeds for large-scale image retrieval. It is obvious that the proposed schemes are much faster than the baselines when distractors are added into the benchmarks. The search time is controlled within one second, and the value is much smaller than that of LSH (whose search time per query is over three seconds) on the large-scale datasets. The search time of the hashing methods still increase linearly with the database volume, even though similarity is measured by fast Hamming distance. While for $ \mathrm{IFC}_{\mathrm{HASH}} $ and $ \mathrm{IFC}_{\mathrm{LSE}} $, their basic framework is inverted tables and only a few images in several Voronoi cells are really compared with a query. That is, search times can be effectively reduced and controlled by changing the codebook size.

(3) $ \mathrm{IFC}_{\mathrm{LSE}} $ is not a stable scheme. As illustrated in Tables \ref{table:unsupervised_small} and \ref{table:unsupervised_large}, its performance is not acceptable in some cases. For example, it only achieves 0.1724 MAP score on Holiday plus MIRFlickr when images are encoded by VGG16 net. Additionally, it is restricted by feature dimensions that should be integral multiples of code length. For this reason we do not illustrate its results of the last four networks in Tables \ref{table:unsupervised_small} and \ref{table:unsupervised_large}. We can see that traditional inverted tables do have limitations for indexing CNN (global) features.

(4) $ \mathrm{IFC}_{\mathrm{HASH}} $ is flexible and its performance is determined by the hashing method. In this scheme, we propose a framework that can employ any kind of hash code, and in the experiments we opt for LSH because no training procedure is needed which is more suitable for unsupervised scenario. Yet, the precision of $ \mathrm{IFC}_{\mathrm{HASH}} $ is mainly affected by hashing and it shows close performance as outlined in Tables \ref{table:unsupervised_small} and \ref{table:unsupervised_large}.

(5) Performance of the CNN networks affects precision of the proposed schemes. To validate adaptability, we perform experiments on different popular CNN networks. Generally, the higher MAP scores for the brute force method brings improvements in terms of accuracy to the other schemes. For example, brute force in AlexNet has higher MAP (0.3185) than VGG19 net (0.2206) on the Oxford dataset. As a result, all the other AlexNet schemes achieve a higher precision than the corresponding VGG19 net schemes. But this rule is not always accurate, as illustrated in Tables \ref{table:unsupervised_small} and \ref{table:unsupervised_large}, since there are many other factors that influence precision. Nevertheless, network performance should be a consideration in improving accuracy.

Another benefit of the proposed schemes is their ability to conserve memory and storage, given they only store image IDs and binary codes (in bits). 

\subsection{Comparison on Supervised Benchmark}

\begin{table}[t]
\caption{Evaluation on NUS-WIDE (supervised benchmark).}
\label{table:supervised}
\begin{center}
\begin{tabular}{|c|c|>{\columncolor{gray!25}}c|c|>{\columncolor{gray!25}}c|}
\hline
\multirow{3}{*}{Method}
&\multicolumn{4}{c|}{Dataset} \\
\cline{2-5}
&\multicolumn{2}{c|}{NUS-WIDE} &\multicolumn{2}{c|}{NUS-WIDE+Flickr1M} \\
\hhline{~----}
& MAP & Time & MAP & Time \\
\hline\hline
\multicolumn{5}{|c|}{AlexNet} \\
\hline
VDSH & 0.5862 & 0.254 & 0.4032 & 1.675 \\
\hline
$ \mathrm{IFC}_{\mathrm{VDSH}} $ & 0.5865 & 0.043 & 0.3812 & 0.096 \\
\hline\hline
\multicolumn{5}{|c|}{VGG16} \\
\hline
VDSH & 0.5893 & 0.305 & 0.3393 & 2.167 \\
\hline
$ \mathrm{IFC}_{\mathrm{VDSH}} $ & 0.5915 & 0.050 & 0.3204 & 0.103 \\
\hline\hline
\multicolumn{5}{|c|}{VGG19} \\
\hline
VDSH & 0.5908 & 0.315 & 0.3615 & 0.919 \\
\hline
$ \mathrm{IFC}_{\mathrm{VDSH}} $ & 0.5911 & 0.059 & 0.3341 & 0.106 \\
\hline\hline
\multicolumn{5}{|c|}{GoogLeNet} \\
\hline
VDSH & 0.5031 & 0.313 & 0.3144 & 1.826 \\
\hline
$ \mathrm{IFC}_{\mathrm{VDSH}} $ & 0.5057 & 0.037 & 0.2912 & 0.090 \\
\hline\hline
\multicolumn{5}{|c|}{ResNet50} \\
\hline
VDSH & 0.5639 & 0.315 & 0.3378 & 2.300 \\
\hline
$ \mathrm{IFC}_{\mathrm{VDSH}} $ & 0.5652 & 0.047 & 0.3159 & 0.098 \\
\hline\hline
\multicolumn{5}{|c|}{ResNet101} \\
\hline
VDSH & 0.5621 & 0.318 & 0.3515 & 2.096 \\
\hline
$ \mathrm{IFC}_{\mathrm{VDSH}} $ & 0.5643 & 0.069 & 0.3308 & 0.105 \\
\hline\hline
\multicolumn{5}{|c|}{ResNet152} \\
\hline
VDSH & 0.5637 & 0.315 & 0.3606 & 2.224 \\
\hline
$ \mathrm{IFC}_{\mathrm{VDSH}} $ & 0.5646 & 0.048 & 0.3421 & 0.098 \\
\hline
\end{tabular}
\end{center}
\end{table}

Recent hashing methods are based on deep learning \cite{xia2014supervised,lin2015deep,lai2015simultaneous,erin2015deep,zhao2015deep,zhang2016ssdh,linlearning,liudeep,zhuang2016fast}, which perform retrieval on supervised datasets. The difference in unsupervised searches is that true matches are judged in terms of whether two images share common labels but not similar content. We employ VDSH \cite{VDSH_zhang2016efficient} to calculate hash codes within $ \mathrm{IFC}_{\mathrm{HASH}} $ to make a specific scheme named $ \mathrm{IFC}_{\mathrm{VDSH}} $. Unlike most deep hashing methods for image retrieval, VDSH takes feature vectors as inputs but does not operate on images directly, which is more suitable for our proposed scheme. We perform comparative experiments on NUS-WIDE dataset between $ \mathrm{IFC}_{\mathrm{VDSH}} $ and VDSH and illustrate the results in Table \ref{table:supervised}.

As opposed to previous experiments, we set $ R = 50 $ in Eq. \ref{equation:ap} to calculate the MAP (often denoted as mAP in other literature). That is, we only consider the top 50 returned images. This is because NUS-WIDE contains many images per tag and each query has many true matches. However, $ \mathrm{IFC}_{\mathrm{HASH}} $ only returns top results, and correct images outside candidate set are not returned. Therefore, it is unfair to calculate MAP score on the whole ground-truth. 

From the results illustrated in Table \ref{table:supervised}, we can draw the following conclusion: 

$ \mathrm{IFC}_{\mathrm{VDSH}} $ is much faster than VDSH at the cost of a little precision. For example, on NUS-WIDE plus Flickr1M datasets, $ \mathrm{IFC}_{\mathrm{VDSH}} $ loses $ 5\% $ MAP score but is $ 17 $ times faster than VDSH. Unlike unsupervised cases, this conclusion is substantially correct when there are no distractors. Because NUS-WIDE is a medium-scale dataset, acceleration of search speed is more obvious than on the other three unsupervised benchmarks. From these experiments, we demonstrate that $ \mathrm{IFC}_{\mathrm{VDSH}} $ fits any hashing method.

\section{Conclusion}
In this paper, a simple but effective indexing framework for CNN features is proposed. Inspired by previous works on the BoW model, we modify inverted tables, making them suitable for accelerating retrieval of global CNN vectors. To compensate for quantization errors, several strategies are fully investigated. We design two schemes, \emph{i.e.}, an initial and an accelerated versions. Experiments on four benchmarks show that the second scheme, \emph{i.e.},  
$ \mathrm{IFC}_{\mathrm{VDSH}} $, greatly improves search speeds with little loss of precision. In future works, we intend to apply the proposed indexing technique to other applications, such as cross-media retrieval, to deal with large-scale scenarios.

\bibliographystyle{spmpsci}
\bibliography{template}

\end{document}